\documentclass[sigconf]{acmart}
\usepackage{amssymb}

\usepackage{latexsym}

\usepackage{url}

\usepackage{multirow}
\usepackage{balance}
\usepackage{amsfonts}
\usepackage{amsmath}

\usepackage{amsthm}
\usepackage{graphicx}

\usepackage{microtype}
\usepackage{url}
\usepackage{enumitem}
\usepackage{wrapfig,lipsum}
\usepackage{multirow}
\usepackage{makecell}
\usepackage{wrapfig}
\usepackage{tabularx}
\usepackage[boxed, ruled,vlined,linesnumbered]{algorithm2e}
\usepackage{algpseudocode}
\usepackage{microtype}
\usepackage{etoolbox}
\usepackage{subcaption}
\usepackage{float}
\usepackage{booktabs}
\usepackage{caption}
\usepackage{blindtext}
\usepackage{amsmath}
\usepackage[linesnumbered,ruled]{algorithm2e}%

\usepackage{pifont}
\usepackage{tabularx,ragged2e}

\usepackage{geometry} \usepackage{array, makecell}%
\setcellgapes{4pt}

\newcommand{\nop}[1]{}

\newcommand{\ie}{\emph{i.e.}}
\newcommand{\xmark}{\ding{56}}%
\newcommand{\CoExpan}{\mbox{\sf Set-CoExpan}\xspace}
\AtBeginDocument{%
  \providecommand\BibTeX{{%
    \normalfont B\kern-0.5em{\scshape i\kern-0.25em b}\kern-0.8em\TeX}}}

\setcopyright{acmcopyright}
\begin{document}

\title{Guiding Corpus-based Set Expansion by Auxiliary Sets Generation and Co-Expansion}

% \author{Ben Trovato$^{1*}$,xx}
% \affiliation{%

% \affiliation{%

% \affiliation{%

\author{Jiaxin Huang$^{1*}$, Yiqing Xie$^{2*}$, Yu Meng$^{1}$, Jiaming Shen$^{1}$,  Yunyi Zhang$^{1}$,  Jiawei Han$^{1}$}
\affiliation{
\institution{$^1$University of Illinois at Urbana-Champaign, IL, USA} 
\institution{$^2$The Hong Kong University of Science and Technology, Hong Kong, China}
\institution{$^{1}$\{jiaxinh3, yumeng5, js2, yzhan238, hanj\}@illinois.edu \ \ \ $^2$yxieal@ust.hk}
}
\thanks{$^*$Equal Contribution.}

\renewcommand{\shortauthors}{Huang et al.}

\begin{abstract}
\nop{ %JH
  The task of corpus-based set expansion aims to induct an extensive set of entities which share the same semantic class with seed entities from a given corpus. It benefits a wide range of downstream applications in knowledge discovery: web search, query suggestion, taxonomy construction and so on. Existing corpus-based set expansion algorithms typically bootstrap the given seeds by incorporating lexical patterns and distributional similarity. However, they suffer from semantic drift caused by expanding the seed set freely without guidance. We propose \CoExpan, a framework that first generates auxiliary sets as closely related to the target set of users' interest, and then perform multiple sets co-expansion that extracts discriminative features by comparing target set with auxiliary sets, to form multiple cohesive sets that are distinctive from one another, thus resolving the semantic drift issue. In this paper we demonstrate that by generating auxiliary sets, we can guide the expansion process of target set to avoid touching those ambiguous areas around the border with auxiliary sets, and we show that our method outperforms strong baseline methods significantly.
}
Given a small set of seed entities (e.g., ``USA'', ``Russia''), corpus-based set expansion is to induce an extensive set of entities which share the same semantic class (\textit{Country} in this example)  from a given corpus. 
Set expansion benefits a wide range of downstream applications in knowledge discovery, such as web search, taxonomy construction, and query suggestion. 
Existing corpus-based set expansion algorithms typically bootstrap the given seeds by incorporating lexical patterns and distributional similarity. 
However, due to no negative sets provided explicitly, these methods suffer from semantic drift caused by expanding the seed set freely without guidance. 
We propose a new framework, \CoExpan, that automatically generates auxiliary sets as negative sets that are closely related to the target set of user's interest, and then performs multiple sets co-expansion that extracts discriminative features by comparing target set with auxiliary sets, to form multiple cohesive sets that are distinctive from one another, thus resolving the semantic drift issue. 
In this paper we demonstrate that by generating auxiliary sets, we can guide the expansion process of target set to avoid touching those ambiguous areas around the border with auxiliary sets, and we show that \CoExpan outperforms strong baseline methods significantly.
\end{abstract}



\copyrightyear{2020}
\acmYear{2020}
\setcopyright{iw3c2w3}
\acmConference[WWW '20]{Proceedings of The Web Conference 2020}{April 20--24, 2020}{Taipei, Taiwan}
\acmBooktitle{Proceedings of The Web Conference 2020 (WWW '20), April 20--24, 2020, Taipei, Taiwan}
\acmPrice{}
\acmDOI{10.1145/3366423.3380284}
\acmISBN{978-1-4503-7023-3/20/04}


\maketitle

%\section{Introduction}
%
%
%ACM's consolidated article template, introduced in 2017, provides a
%consistent \LaTeX\ style for use across ACM publications, and
%incorporates accessibility and metadata-extraction functionality
%necessary for future Digital Library endeavors. Numerous ACM and
%SIG-specific \LaTeX\ templates have been examined, and their unique
%features incorporated into this single new template.
%
%If you are new to publishing with ACM, this document is a valuable
%guide to the process of preparing your work for publication. If you
%have published with ACM before, this document provides insight and
%instruction into more recent changes to the article template.
%
%The ``\verb|acmart|'' document class can be used to prepare articles
%for any ACM publication --- conference or journal, and for any stage
%of publication, from review to final ``camera-ready'' copy, to the
%author's own version, with {\itshape very} few changes to the source.
\section{Introduction}

The task of set expansion is to enrich the set of given seeds belonging to a certain semantic class (e.g., given ``\textit{United States}'', ``\textit{Canada}'', and ``\textit{China}'', set expansion algorithms would extend the set by terms such as ``\textit{France}'', ``\textit{Germany}'' and ``\textit{Australia}'', which also belong to the semantic class of \textit{Country}). Set expansion benefits a variety of downstream applications in natural language processing and knowledge discovery, such as web search \cite{Chen2016LongtailVD}, query suggestion \cite{Cao2008ContextawareQS}, question answering \cite{Prager2004QuestionAU,Wang2008AutomaticSE}, and taxonomy construction \cite{Velardi2013OntoLearnRA}.

\begin{figure}[t]
\centering
\includegraphics[width=0.8\linewidth]{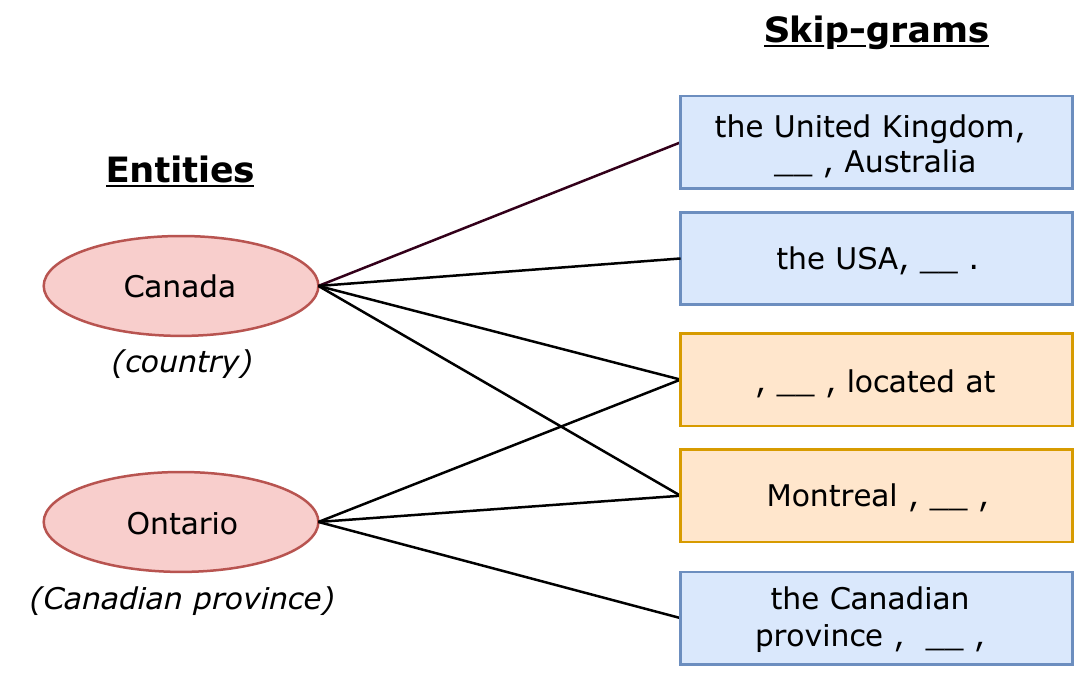}
\caption{Examples of shared skip-grams between country and province/states entities.}
\label{fig:ssg}
\vspace{-0.3cm}
\end{figure}

Existing set expansion methods that rely on a large corpus typically bootstrap the initial seeds by refining the context feature pool and candidate term pool in an iterative manner. Though achieving reasonably good results, they still suffer from the problems of semantic drifting and entity intrusion in the expansion process. Typical errors come from closely related entities of different granularity or semantic types. Consider the example in Figure \ref{fig:ssg}, we draw a bipartite graph where entities are linked to local contexts they appear in. Despite the difference in granularity of ``Canada'' as a country and ``Ontario'' as a Canadian province, they share some local contexts. These shared contexts, when introduced into the context feature pool, result in cross-category expansion and harm the final output continuously in the iterative process. 
Similarly, the shared context may bring in alien semantic types in set expansion.
For example, a disease ``\textit{lung cancer}'' may bring in symptom ``\textit{chest pain}'' since both may share some common local contexts, such as ``\textit{suffer from $\_\_$}''.
\nop{ %JH rewritten
Another example shows how entities that are related but of different semantic type can be expanded: Suppose we want to expand entities like ``\textit{lung cancer}'' from the class of \textit{Diseases}. Terms such as ``heart attack'' are only \textit{Symptoms} instead of \textit{Diseases}, but both words can fit into the local context of ``\textit{suffer from $\_\_$}'', thus entities from closely related concepts are very likely to interfere with the expansion process of the user's target class. 
}
Such kind of semantic drift is caused by the uncontrolled expansion without guidance. 
% Without the overview of the whole semantic class, it is possible that initial seeds lie on the border of that class. 
Since there are no negative sets explicitly given, and either initial seed or generated new members could be at the border of multiple sets,
uncontrolled expansion will likely cross the border and generate off-class entities. 

Some previous studies have explored to incorporate external knowledge in set expansion using either implicit supervision from other queries or human given negative examples. 
\cite{Lin2003BootstrappedLO,Thelen2002ABM} use other queries in the dataset to provide mutual exclusive signals, relying on the assumption that different queries belong to different semantic classes, which does not necessarily hold when other queries are not visible. 
\cite{Jindal2011LearningFN} utilizes human given negative examples to constrain the scope of set expansion. 
However, an implicit assumption is that possible directions of error need to be predicted in advance, and human effort is needed to provide cases for each direction. 
The applicability of these methods is hindered by their prerequisites as they are not adaptable to specialized domains with very few experts.

To tackle the problem of semantic drift in set expansion, we explore a fully automated approach without negative sets or assumptions explicitly provided by human experts.
We observe that a typical source of error comes from the entities from different semantic classes that may share some common relations to the target class. 
Word embedding has shown effective in capturing certain relations between entities, such as the parallelism of $v(Berlin) - v(Germany)$ and $v(Paris) - v(France)$ in the embedding space, which can also be used to distinguish \textit{City} from \textit{Country}.
% commonly captures the relation of "\$Captial of \$City".  
If one can recognize such subtle relationships between entities belonging to different semantic classes, one can capture such subtlety and use it to separate entities from different classes and conduct set co-expansion.
Such 
% If these closely related but quite confusing classes are recognized in the first place, we can conduct 
co-expansion may incorporate signals from all the related, participating classes, keep warning the target class not to cross over the boundaries of its possible rivals, and guide the expanding direction of each set by avoiding to bump into each other's territory. 
The co-expansion of such multiple rival sets benefits each other from mutually exclusive signals, and the quality of multiple sets expansion can be improved simultaneously.

\begin{figure*}[t]
\centering
\includegraphics[width=0.95\linewidth]{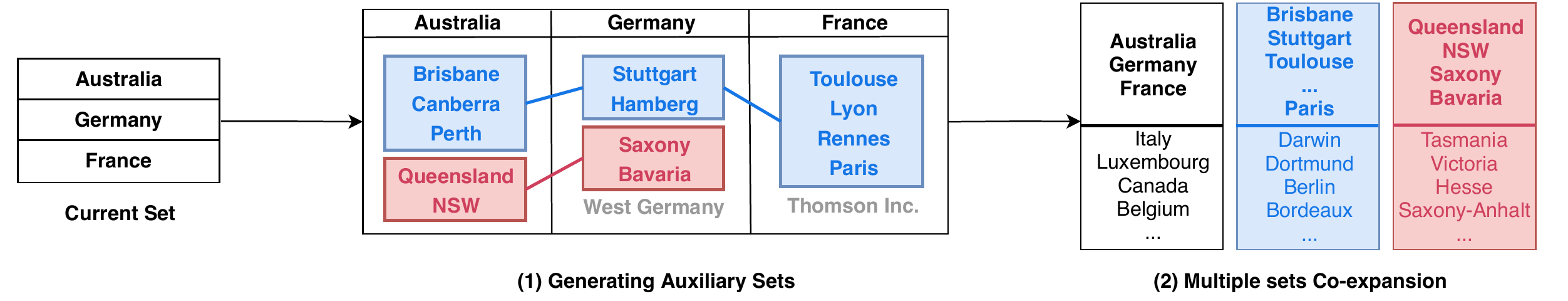}
\caption{Workflow of \CoExpan in one iteration: For user-input country names, related terms such as provinces and cities are retrieved and clustered into auxiliary sets. Multiple sets are then co-expanded by extracting discriminative context features. }
\label{fig:frm}	
\vspace{-0.3cm}
\end{figure*}

We propose a framework called \CoExpan with its workflow shown in Figure \ref{fig:frm}. 
It 
% first generates auxiliary sets which hold certain subtle relations in the embedding space with the user-interested semantic class in an unsupervised way, and then extracts the most discriminative features to tell the target class from auxiliary sets.
% Our framework 
consists of two modules that are operated iteratively: Auxiliary sets generation module that finds auxiliary sets holding certain relations with the target set in an unsupervised way, and multiple sets co-expansion module that takes multiple sets as input and extracts the most discriminative features to tell the target class from auxiliary sets. The auxiliary sets generation module first retrieves semantically related terms to each seed element in an embedding space that captures topical similarity. These related terms are then grouped by their semantic types, captured unsupervisedly by intra-seed clustering and inter-seed merging. The latter step captures parallel relations among inter-seed terms by imposing a parallelogram in the embedding space. The second module, multiple sets co-expansion, takes the target seed set as well as auxiliary sets as input. By incorporating knowledge from both seed set and auxiliary sets, we can control the expanding directions of multiple sets. Specifically, context features are scored by how well they can tell different sets apart, and the algorithm drives the expanding direction away from ambiguous areas. 

We demonstrate the effectiveness of our framework through a series of experiments on two real-world datasets, and show that \CoExpan outperforms strong baselines significantly. We also provide examples and qualitative analysis of auxiliary sets to showcase the effectiveness and generalizability of the proposed approach.
% explain why and how they consistently benefit set expansion given queries of various semantic classes.

Our contribution can be summarized as follows: (1) An auxiliary sets generation module that extracts entities that share certain relations to thus different from the target semantic class. (2) A multiple sets co-expansion module that utilizes the mutual exclusive signals to extract features with the most discriminative power to expand multiple sets simultaneously. (3) Comprehensive experiments and case studies that prove the effectiveness of \CoExpan and justify its generalizability.

\vspace{-0.1cm}

% \vspace{-0.2cm}
\section{related work}

% web-based methods
% google set, SEAL and Lyretail
\smallskip
\noindent \textbf{Web-based Set Expansion.}~
Most of early set expansion methods are developed in online web applications.
They typically submit a query consisting of seed entities to a web search engine and extract entities from retrieved web pages.
Google Sets \cite{googleSet}, now discontinued, uses latent semantic indexing technique to web pages for calculating word similarities. 
SEAL \cite{Wang2008AutomaticSE} extracts item lists from retrieved pages and constructs a heterogeneous network on which a graph-based ranking model is applied to rank entities.  
Lyretail \cite{Chen2016LongtailVD} develops a supervised page-specific extractor to extract long-tail entities from the web. 
All these methods require an external search engine for online data collection, which is quite time-consuming.
Our paper does not focus on this type of setting that uses retrieved documents as information resource. Instead, we address corpus-based set expansion problems and the related work is introduced below.
% Therefore, recent studies focus more on expanding entity sets by utilizing a large corpus as the information resource.

% Early term set expansion methods are developed mostly in web applications, and this line of work uses a search engine to retrieve webpages by queries consisting of seed entities. Google Sets \cite{googleSet}, now discontinued, is one of the earliest work, which uses latent semantic indexing to pre-compute similar words.
% %It automatically find the lists containing similar words from the web and analyzes the obtained webpages. 
% Later, SEAL \cite{} extracts lists of items from retrieved documents, and constructs a heterogeneous network and use a graph-based ranking model to rank the terms.  
% Lyretail \cite{Chen2016Lyretail} deals with the expansion of long-tail terms by building a supervised webpage-specific extractor. 
% All these methods use an external search engines to collect data from online resources and is quite time-consuming, thus recent studies focus more on utilizing an existing large corpus as information resource.

% corpus-based methods 
% setExpander and CaSE
% different focus
% egoset and singleton
% \subsection{Corpus-based Set Expansion}
% \subsubsection{Single query based Set Expansion}
\smallskip
\noindent \textbf{Corpus-based Set Expansion using Single Query.}~
Many corpus-based set expansion methods \cite{Etzioni2005UnsupervisedNE,Gupta2014ResearchAA,Gupta2014ImprovedPL,Shen2017SetExpanCS,Mamou2018setExpander,Zhu2019FUSEMS} are developed to bootstrap a \emph{single seed set query} by iteratively refining a context feature pool and a candidate entity pool. 
Given the context feature pool, these methods build the candidate entity pool by adding only entities that co-occur frequently with high-quality context features.
Meanwhile, they refine the context feature pool by including only those features which are commonly shared by entities in the expanded set. 
Based on this philosophy, SetExpan \cite{Shen2017SetExpanCS} develops a context feature selection module to select quality skip-gram features and designs a rank ensemble module to select quality entities. 
Similarly, SetExpander \cite{Mamou2018setExpander} captures distributional similarity on five different context types and learns a classifier to combine multiple contexts using an additional labeled dataset.
Some corpus-based set expansion methods are designed to focus on one particular aspect of expansion process. 
For example, CaSE \cite{Yu2019CaSE} pursues the expansion efficiency and thus develops a one-time ranking scheme based on a scoring function that incorporates both lexical patterns and distributional similarity. 
EgoSet \cite{Rong2016EgoSetEW} and FUSE \cite{Zhu2019FUSEMS} targets the multi-faceted expansion which allows for each seed term to generate multiple sets belonging to different senses.
Atzori \emph{et al.}, \cite{Atzori2018SingletonSE} focus on singleton expansion (\emph{i.e.}, only one seed in the initial seed set).

Above methods only pay attention to the entities in the user-input seed query, suffering from possible semantic drift to other related classes while expanding the seed set. 
Vyas and Pantel \cite{Vyas2009SemiAutomaticES} propose a human-in-the-loop set refinement method that enables human annotators to identify wrongly expanded entities in each iteration and then removes those context features shared by both positive and negative entities. 
SetExpander \cite{Mamou2018setExpander} implements a system that allows users to validate the correctness of expanded entities and do re-expansion at each iteration. 
These methods require additional human efforts and thus are not directly comparable to our method. 
\smallskip
\noindent \textbf{Corpus-based Set Expansion using Hybrid Query.}~
Other methods proposed to use multiple different semantic classes as negative signals to constrain the scope of sets to be expanded. Basilisk \cite{Thelen2002ABM} expands six categories given in the dataset simultaneously to avoid claiming terms in multiple categories. NOMEN \cite{Lin2003BootstrappedLO} also allows for expanding multiple semantic classes simultaneously, and treat seeds from other classes as negative examples. These methods rely on other implicit queries in the dataset, which may not be relevant enough to provide mutual exclusive signals, and other queries may not even exist on every occasion. What's more, the assumption that different queries belong to different semantic classes may not even holds. \cite{Jindal2011LearningFN} proposes an inference-based algorithm that leverages human-given negative examples to define the semantic class. For example, human can provide the expansion of female tennis players with male tennis players as negative signals. However, the method requires prior knowledge of possible errors among multiple dimensions. \emph{E.g.}, female football players and female volleyball players can also serve as negative examples and the list is never exhausted.
 
 These refining methods basically proves the idea that expanding multiple semantic classes simultaneously leads to better results, since  mutual exclusive signals can help distinguish those ambiguous terms. However, they all rely on external supervision from other classes provided by human, thus are not comparable to our method.

% setExpan
%The current paper is the extension of SetExpan, proposed by Shen et al \cite{Shen2017SetExpanCS}. It uses an iterative pattern-based bootstrapping framework. In each iteration, it first selects context features based on both initial seeds and expanded terms, then rank entity candidates based on these relevant features. However, setExpan may expand entities that belong to relevant but not the same semantic classes. In comparison, our approach expands several relevant seed sets at the same time, and selects context features that can distinguish these sets. Our experiments show that the expansion of relevant seed sets can mutually enhance each other by reducing this kind or mistakes.

% \vspace{-0.2cm}
\section{Problem Definition}
In this section we give the definition of the task of corpus-based set expansion.
 The inputs are a collection of documents $\mathcal{D}=\{d_1,d_2,...,d_{|\mathcal{D}|}\}$ and a seed set $S$ provided by user. The elements in $S$ are seed terms that belong to the same target semantic class $C_{T}$, which is a superset of $S$. The output is a ranking list of entities which is computed based on the lexical context features and distributed representations learned from the given corpus.
 
In our method, we will not only look into the given set of seeds $S$, but also try to extract other seeds forming multiple auxiliary sets $C_{aux}=\{S_1\subsetneq C_1, S_2\subsetneq C_2, S_3\subsetneq C_3, ..., S_{|P|}\subsetneq C_{m}\}$ as rival sets that are closely related to but different from the target semantic class $C_{t}$, to guide the expanding process by incorporating the mutual exclusive knowledge from target and multiple auxiliary sets. Our method automatically generates multiple auxiliary sets and does not require any human effort.

\section{Methodology}

In this section, we introduce our proposed method by first giving an overview in Section \ref{sec:overview} and then describing the details of two modules in Section \ref{sec:gen} and \ref{sec:exp}.

\vspace{-0.25cm}
\subsection{Method Overview} \label{sec:overview}
For a given set of query entities, we gradually expand the seed set by adding a fixed number of new entities at each iteration. Two modules collaborate with each other in an iterative manner: Auxiliary Sets Generation and Multiple Sets Co-Expansion. 

\begin{figure}[t]
\vspace{-0.3cm}
\centering   
\includegraphics[width=0.98\linewidth]{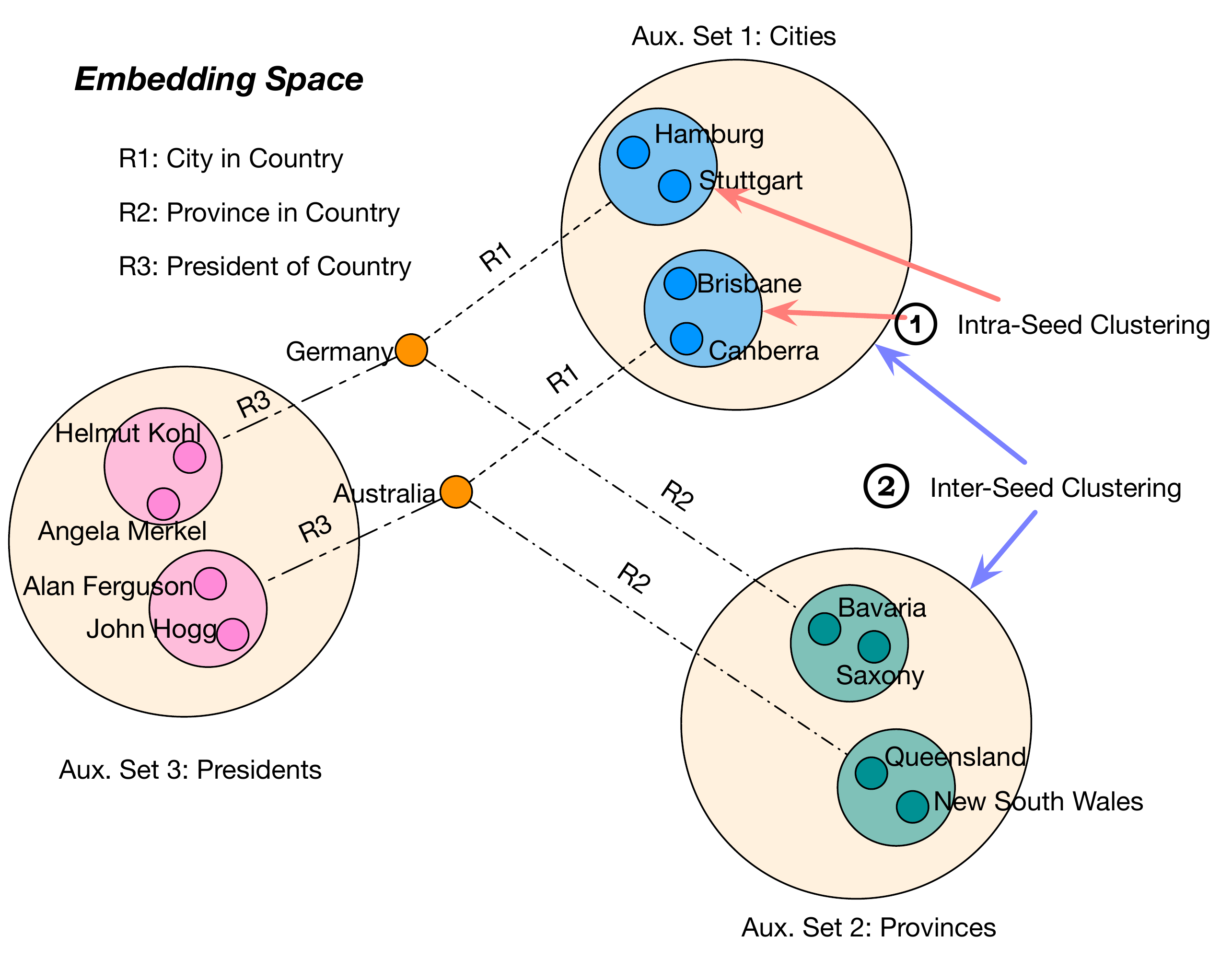}
\caption{Auxiliary sets generation: Related terms of each seed are first clustered according to their semantic types; Cross-seed auxiliary sets are formed by capturing parallel offsets in the embedding space.}
\label{fig:merge}
\vspace{-0.3cm}
\end{figure}

The first module is pictured in Figure \ref{fig:merge}. For all seeds in the target set, we automatically generate auxiliary sets as rival classes, and auxiliary set entities share the same relations with the target set. In this figure, the user wants to expand country names, and \CoExpan can extract entities that share certain relations with given seeds (e.g., \textit{Brisbane} as a city located in \textit{Australia} as a country, and \textit{Merkel} as the prime minister of \textit{Germany} as a country), and thus forming auxiliary sets of \textit{``Cities''}, \textit{``Provinces''} and \textit{``Presidents/Prime Ministers''}. These new semantic classes benefit the model in distinguishing different granularities of the same concept or different semantic types. 

In the second module, given target seed set and its auxiliary sets, we design a multiple sets co-expansion algorithm to let all the sets expand simultaneously in recognition of the existence of other sets, and treat elements in other sets as pitfalls to be avoided. Specifically, we use a context dependent similarity measure to describe the similarity of pairs of entities. We then extract the context features that maximize the similarity of pairs of elements from the same set, and minimize the similarity of pairs of elements from different sets. In this way, we can guide the set expansion process to avoid touching the ambiguous skip-grams shared by different semantic classes, then the target set and auxiliary sets can gradually expand their territories without grabbing off-class entities. 

\vspace{-0.25cm}
\subsection{Auxiliary Sets Generation}
\label{sec:gen}
As shown in Figure \ref{fig:merge}, in order to generate auxiliary sets that contain entities closely related to but different from the target semantic class, we aim to find entities forming groups that hold certain relations with given seed entities. For example, ``\textit{Hamburg}'' and ``\textit{Stuttgart}'' can form a small group in the embedding space due to their proximity in semantics. Another small group consists of ``\textit{Brisbane}'' and ``\textit{Canberra}'', and both groups share a parallel relation with seeds ``\textit{Germany}'' and ``\textit{Australia}'', represented as
\vspace*{-0.1cm}
\begin{equation}\label{eq:rel}
    Relation(e_1\in C_T, g_1) \approx Relation (e_2 \in C_T, g_2)
\vspace*{-0.1cm}
\end{equation}

Using the information that $e_1$ and $e_2$ belong to the same semantic type, we can infer that two groups $g_1$ and $g_2$ are of the same semantic type and merge them into a cross-seed group. It is obvious to see that these cross-seed groups ensure consistent relations with the seed elements in the target semantic class, so that they can serve as auxiliary sets that we aim to look for.

In the following we will introduce how we first retrieve pools of related words to seed elements, and then perform intra-seed and inter-seed clustering to form these auxiliary sets.

\subsubsection{Semantic Learning and Related Terms Retrieval for Seed Elements}\label{sec:emb}

Word2Vec \cite{Mikolov2013DistributedRO} has achieved wide success in capturing word semantics. To retrieve relevant terms to a certain seed, it is common practice to search for its nearest embedding space neighbors. Word2Vec assumes semantically similar words have similar local context (words in a fixed window size around the center word), and its corresponding training objective is 
{\small
\begin{align*}
L_l &= \sum_{d\in D} \sum_{1\leq i\leq |d|} \sum_{0<|j-i|\leq h} \log P(w_j|w_i) \\
&= \sum_{d\in D} \sum_{1\leq i\leq |d|} \sum_{0<|j-i|\leq h} \log \frac{\exp{\left(v_{w_i} u_{w_j}\right)} }{\sum_{w_{j'} \in V }\exp{  \left(v_{w_i} u_{w_{j^{'}}}\right) }}
\end{align*}
}where h is the window size. A recent study \cite{Meng2020DiscriminativeTM} shows that document-level co-occurrences of words (\ie, \emph{global contexts}) can capture topical relatedness between words, with the assumption that words appearing in similar documents tend to be topically similar. Here, we employ the same global context loss as in \cite{Meng2020DiscriminativeTM}: 
{\small
\begin{align*}
L_g &= \sum_{d\in D} \sum_{1\leq i\leq |d|} \log P(d|w_i)\\
&= \sum_{d\in D} \sum_{1\leq i\leq |d|} \log \frac{\exp{\left(v_{w_i} u_d\right)}}{\sum_{d^{'} \in D } \exp{\left(v_{w_i} u_{d^{'}}\right)} }    
\end{align*}
}Global contexts characterize document-level co-occurrence statistics, while local contexts capture word-level co-occurrence statistics within local context windows. To encode the information of both sets of contexts into word embeddings, we propose an overall training objective with a weighted sum of global and local context losses.
$$L=L_l + \lambda L_g$$

% the most relevant words for each seed element is retrieved at the last epoch of embedding training.
% At the end of each epoch in embedding training, 
After the embedding training is finished, for each entity $e$, we use cosine similarity to retrieve the $k$ most related terms, referred as $r_e$, from the vocabulary.
% for each entity $e'$ in the vocabulary and get the k highest ones, denoted as $Z_e$.
% $$Sim(e',e) = \cos (w_{e'},w_{e}) $$

%Since we want to form cross-seed parallel relations, the retrieved candidates should be largely relevant to one of the seed entities, and have minor relevance with the other. Therefore, it is crucial to learn an embedding space that distinguishes words that are only relevant to one seed entity. 
% We refine the embedding space by adding a regularization term to push words in $Z_e$ toward their relevant seed entity $e$ while away from the other ones in each epoch. Let $\overrightarrow{E}$ be a list of entities in the target seed set, $\overrightarrow{E} = [e_1, e_2, \cdots, e_{|S|}]$, $e_i \in S$.
% $$L_{r} = \sum_{e\in S} \sum_{w\in Z_e} KL(\mathbb{I}[w \in Z_{\overrightarrow{E}}] \|Sim(w, \overrightarrow{E})) $$
% By minimizing the above KL-divergence, the most similar words are trained to be closer to their relevant entities, and will gradually impact other similar words to be clustered around the relevant entity in later iterations of embedding training.

\vspace{-0.3cm}
\subsubsection{Cross-Seed Parallel Relations Clustering}

Figure \ref{fig:merge} shows an example of how auxiliary sets are confirmed as different semantic classes by the constraint of cross-seed parallel relations. Specifically, for a seed country ``Australia'', its related words $r_{Australia}$ can first be clustered into small initial groups of the same semantic type: $G_{Australia}=\{g_{Australia}^1, g_{Australia}^2\}$, where $g_{Australia}^1=\{\text{Brisbane, Canberra, Perth}\}$ and $g_{Australia}^2=\{\text{Queensland, NSW}\}$. For other seed elements like ``Germany'', several groups can also be clustered to form $G_{Germany}$. 
Then inter-seed clustering can be performed to merge $g_{Australia}^1$ and $g_{Germany}^1$ into an auxiliary set since they hold parallel relations(City in Country) with corresponding seed element. 

% Then the groups with same cross-entity relations will be detected as $g_{Australia}^1 =\{\text{Brisbane, Canberra, Perth}\}$ and $g_{Germany}^1 =\{\text{Hamburg, Stuggart}\}$ are both cities in their subscript countries. Then a new parallel set could be confirmed by merging $g_{Australia}^1$ and $g_{Germany}^1$ to $S_1$, where $S_1$ is the subset of $C_1$, a semantic class of cities.

% There are two challenges in detecting these groups and relations: (1) It is hard to derive the initial clusters like $g_{Australia}^1 =\{\text{Brisbane, Canberra, Perth}\}$ in very high precision, since this step does not rely on any typing information and is totally unsupervised. (2) Cross-seed parallel relations are also derived in an unsupervised way. 

% We overcome these difficulties by performing a two-step procedure: initial group clustering and merging cross-seed groups.

We perform two steps in extracting these auxiliary sets: (1) Intra-seed clustering that clusters terms in each $r_e$ into initial groups of the same semantic class $g_e^i$ in a high precision. (2) Inter-seed clustering that merges initial groups that share same relations with corresponding elements into auxiliary sets.

In the first step, we cluster the retrieved related words in $r_e$ by their proximity in the embedding space. 
Since 
% these related words could be of any semantic class, and 
we obviously do not have a prior knowledge of the number of classes, 
% thus it would be wiser to do hierarchical clustering on these words. What's more, 
% not all the terms share common semantic class with other terms, which means
and some outlier terms might not even be included in any high quality clusters, we apply Hierarchical Agglomerative Clustering (HAC) \cite{Rokach2005ClusteringM}, a bottom-up hierarchical clustering method.
% which has a potential for us to derive initial clusters with a high precision. 
The original result of HAC on $n$ data points is a binary tree-based representation named dendrogram with $n$ leaf nodes and $n-1$ internal nodes. A problem of directly applying this clustering would be the level chosen for splitting the data points. 
As groups of terms would be mixed with different semantic classes if the level chosen is too high, or missing meaningful groups if the level chosen is too low. 
% stopping criterion of merging small groups together. 
To solve this issue, we insert user input seed set $S$ into the terms to be clustered, and the philosophy is that we stop the merging process when a seed element from target class is about to be merged with element related words. 
% Formally, we get the final clustering result at maximal level $l$ (${l}\leq |r_e-1|$) which satisfies:
% $$|Children(l) \cap r_e | \cdot |Children(l) \cap S | > 0 $$
% The key idea is to add a cannot-link so that when a retrieved related term and a seed entity is about to be merged into a new group, the new group might be a mixture of terms from different classes. Therefore, the merging process stops and the clusters formed at the current iteration which have more than one entity become our initial clusters. 
% where $Children(l)$ refers to all the children of the only internal node at the $l^{th}$ level.
In this way, we make sure that the new clusters generated are of different semantic class from the original seed entities that belong to $C_T$.

In this HAC, our similarity measure uses the Euclidean distance between terms in the BERT \cite{Devlin2019BERTPO} embedding space, where the representation of each entity is calculated by averaging the contextualized embedding of each occurrence in the corpus. We choose \textit{``complete''} linkage for constructing more compact clusters. 

% In order to merge together initial clusters of same semantic class like $g_{Australia}^1 =\{\text{Brisbane, Canberra, Perth}\}$ and $g_{Germany}^1 =\{\text{Hamburg, Stuggart}\}$, we hypothesize that if two clusters have the same relations with their corresponding seed entities that belong to the same semantic class $C_T$ (\textit{Australia} and \textit{Germany} respectively in this case), then these two clusters also belong to a same semantic class $C_i \in P$.

In the second step, we use inter-seed clustering to capture parallel relations in Equation~\eqref{eq:rel}, and we interpret the Relation function as linear operations. Word analogy examples in \cite{Mikolov2013DistributedRO} shows the Word2Vec embedding space hold the property of linearly capturing relations between words, so that $v(Berlin) - v(Germany) \approx v(Paris) - v(France)$, with recent theoretical explanation in \cite{Allen2019AnalogiesET}. 
Studies on relation extraction \cite{Bordes2013TranslatingEF, Wang2014KnowledgeGE, Lin2015LearningEA} relies on the idea that for a triplet of head <entity, relation, tail entity> and their vector representation <$h, r, t$>, the linear relation holds that $h+r\approx t$.
% This equation basically shows that if two pairs of entities have same relations, then their difference in the vector space are parallel. The theoretical explanation is also given in \cite{Allen2019AnalogiesET}.

We also apply the above idea to capture cross-seed parallel relations. Specifically, we select the $i^{th}$ initial group of element $e$, denoted as $g_{e}^{i}$. We then consider whether it can be merged with $g_{e'}^{j}$ of a different seed element. To enlarge $g_{e}^{i}$ to cover more terms of the same type, we first expand $g_{e}^{i}$ into a larger group $g_{e,expanded}^{i}$ by a single query version of our set expansion algorithm later introduced in section \ref{sec:exp}. 
% of at least 15 entities.
To represent the center of this expanded group, we take the average over all elements in $g_{e,expanded}^{i}$. 
% We denote $v_w$ to be the vector representation of word $w$ in the Word2Vec embedding space.
$$v_{e,expanded}^{i} = \frac{\sum_{w\in g_{e,expanded}^{i}} v_w}{|g_{e,expanded}^{i}|}$$
After that, we perform a parallel translation for $g_e^i$ from element $e$ to $e'$ and obtain a pseudo center point for the new element $e'$, 
$$v_{e\rightarrow e'}^{i} = v_{e'} - v_e + v_{e,expanded}^{i}$$
We then retrieve 15 nearest neighbors from $v_{e'}$ as the pseudo transitioned group $g_{e\rightarrow e'}^{i}$ 
% and count their overlap with $g_{e',expanded}^{j}$. 
We then try to find an initial group $g_{e'}^{j}$ of element $e'$ that satisfies
$$g_{e\rightarrow e'}^{i} \cap g_{e'}^{j} > \eta $$
where $\eta$ is the merging threshold that determines whether two groups have enough overlapping terms.
After conducting this merging process, 
% we only retain those initial clusters that are merged with clusters under different seed entities. 
we remove those unmerged initial groups to ensure that each remaining cluster ensures strong relations with seed elements by owning initial groups that share parallel cross-seed relations. We then treat each of these larger groups as an auxiliary set to distinguish our target seed set from. Algorithm \ref{alg:aux} outlines this whole process of parallel relations clustering.
% across different entities in the seed set $S$, thus belong to a new semantic classes that holds a robust relation with the original semantic class $C_T$.

% By now we have generated seeds for multiple parallel sets that hold strong relations with the original input set $S$. These strong relations ensure that they are related but different from the original semantic class. We will show how to utilize seeds from these sets to benefit our model in distinguishing between these different semantic classes.

\small
\vspace{-0.3cm}
\begin{algorithm}
    \SetKwInOut{Input}{Input}
    \SetKwInOut{Output}{Output}

    \Input{Initial seed set $S$; merging threshold $\eta$;
    \newline Retrieved related terms for each seed element $\{r_e | e \in S \} $}
    \Output{Auxiliary sets $C_{aux}$}
    \For{$e$ in $S$}
      {
        $G_e \gets $ Agglomerative Clustering($C_e \cup \{e\}$, 
        \newline stop when $\exists g \in G_e, g \cap S$ * $|g \cap C_e| > 0$)
        
        $G_e \gets \{ g \in G_e | |g| \geq 2 \}$
        
      }
      
    $G \gets \bigcup\limits_e G_e $
      
    \For{ $e, e'$ in S }
    {
    	\For{ $g_e^i \in G_e$, $g_{e'}^j \in G_{e'} $ }
    	{
    		$g_{e,expanded}^i \gets $ expand( $g_e^i$ )
    		
    		$g_{e',expanded}^j \gets $ expand( $g_{e'}^j$ )
    		
    		$v_{e,expanded}^i \gets \frac{\sum_{w\in g_{e,expanded}^i} v_w}{|g_{e,expanded}^i|}$
    		
    		$v_{e\rightarrow e'}^{i} \gets v_{e'} - v_e + v_{e,expanded}^i$
    		
    		$g_{e\rightarrow e'}^{i} \gets $  15 nearest entities around $v_{e\rightarrow e'}^{i}$ 
    		
    		\If{ $|g_{e\rightarrow e'}^{i} \cap g_{e',expansion}^j | \geq \eta $ }
    		{
    			merge  $g_e^i$ and $g_{e'}^j$ in $G$
    		}
    		
    	}
    
    }
    
    $C_{aux} \gets \{ g \in G | |g| \geq 2 \}$
    
    Return $C_{aux}$ 
     
    \caption{Cross-Seed Parallel Relations Clustering}
    \label{alg:aux}
\end{algorithm}
\normalsize
% \vspace{-0.3cm}

\vspace{-0.4cm}
\subsection{Multiple Sets Co-Expansion}
\label{sec:exp}

\subsubsection{Overview of the module}\label{sec:single_exp}

We iteratively refine two key elements in set expansion: feature pool and candidate pool. Feature pool stores common context features of seed entities, which best describe the target semantic class. Candidate pool stores the possible candidate entities to be expanded, and they are narrowed down by co-occurrence with features in the feature pool.

In each iteration, candidate terms that are most similar with the current expanded set are expanded.
When more entities are added into the seed set, the feature pool will be refined by re-selecting the context features that maximize the similarity of user given and currently expanded terms.

We use skip-grams (e.g., $w_{-2}w_{-1}\_w_1w_2$) as local context features that describes an entity mention, and forms a bipartite graph as shown in Figure \ref{fig:ssg}. To derive the weight of the edge connecting an entity with a skip-gram context, we first create a frequency matrix $\Phi_{e,c}\in \mathbb{R}^{N\times M}$ that counts their co-occurrence, where $N$ and $M$ are the number of candidate entities and skip-grams respectively. Then we define the weight connecting entity $e$ and context $c$ as
%  Weights of connecting an entity and a pattern include TF-IDF transformation \cite{}, PMI \cite{}, discounted PMI \cite{}, BM-25 scoring \cite{} and etc. In our model, we inherit \cite{} in choosing TF-IDF transformation as the skip-gram weights, specifically, the score of a skip-gram to an entity is defined as 
$$f_{e,c}=\log (1+\Phi_{e,c})[\frac{\log |N| }{\log \Phi_{e',c}}]$$
which is the TF-IDF transformation \cite{Rong2016EgoSetEW}.
%  This score function resembles TF-IDF in that entities can be seen as \textit{"documents"}, while skip-grams are the \textit{"terms"} that they contain.
An entity $e$ can thus be represented by an $M$-dim vector to derive similarity with other entities.
However, such representation harms set expansion in two aspects: (1) the dimension of M is obviously large and even if we only consider non-zero terms in seed elements, that dimension still grows as the size of expanded set grows. (2) Not all local contexts are type-indicative even if they do have a strong connection with a single seed entity. The weak supervision of given seed elements belonging to the same type can provide valuable information that the shared skip-grams among elements are more type-indicative. Therefore, context dependent similarity benefits set expansion tasks in that it only captures the type-indicative features of entities.

We adopt the context dependent similarity function $Sim(e_i, e_j|F)$ defined in \cite{Shen2017SetExpanCS} using the weighted Jaccard similarity measure:
\begin{equation}\label{eq:sim}
    Sim(e_1,e_2|F)=\frac{\sum_{f\in F}\min(f_{e_1,c},f_{e_2,c}) }{\sum_{f\in F}\max(f_{e_1,c},f_{e_2,c})}
\end{equation}

so that the selected features serve as references to compute the similarity of two entities.

% Despite the effectiveness of these bootstrapping methods, they expand the original seed set freely without guidance, and the expansion process might bump into borders with neighbor semantic classes. However, if multiple related sets are given  together, the expansion can be guided to avoid touching those ambiguous area around the borders of different semantic classes.

\subsubsection{Capturing Contrastive Skip-gram Contexts for Multiple Sets Co-Expansion}

Let us consider the skip-gram context ``the United Kingdom, $\_\_$ , Australia'' in Figure \ref{fig:ssg}, it can be connected to many country entities in the corpus, while very unlikely to be assigned to any other types of entities like provinces or cities. Thus, this kind of skip-grams are discriminative enough in indicating the type of entities. Nevertheless, if we look at the skip-gram ``Montreal , $\_\_$ ,'', it is connected by both ``Canada'' as a country and ``Ontario'' as a Canadian province. The ambiguity of this skip-gram limits its power to discriminate upon different granularities of locations, thus is not helpful in distinguishing related but different sets, and may even cause semantic drift in later iterations. 

In order to capture contrastive skip-gram contexts for multiple sets, skip-grams that make each set cohesive while distinguishing different sets are encouraged. Following this principle, we score skip-grams shared by more entity pairs from the same set higher, while penalize skip-grams shared by arbitrary pairs from different sets, as shown in the following equation.
\vspace{-0.1cm}
\begin{align}\label{eq:cf_sel}
\begin{split}
\small
F^* &=\arg \max_{|F|=Q} \frac{2}{|S|*(|S|-1)} \sum_{e_i,e_j\in S}   Sim(e_i, e_j|F)\\
&- \sum_{S_k,S_{k'}\in C_{aux}} \frac{1}{|S_k|*|S_{k'}|} \sum_{e_i\in S_k, e_j \in S_{k'}}  Sim(e_i, e_j|F) \\
&+ \sum_{S_k \in C_{aux}} \frac{2}{|S_k|*(|S_k|-1)} \sum_{e_i,e_j\in S_k}   Sim(e_i, e_j|F)
\end{split}
\end{align}
where Q is the size of skip-grams to be selected.
To solve the NP-hard optimization problem in Equation \eqref{eq:cf_sel}, we apply a greedy selection process that selects the feature that increments the score the most at each time. This would benefit us in selecting  complimentary features instead of focusing on strong but redundant signals.

% The above criterion chooses skip-gram contexts that force satellite sets as well as the core set to be coherent, so that each of them will expand in a right direction that avoid penetrating each other's territories. 

\subsubsection{Dynamically Adjusting Skip-gram Contexts}\label{sec:flexgram}

Since skip-gram contexts impose hard pattern matching constraints, 
the candidate pool might suffer from a low coverage since 
some infrequent entities might not even appear in those type-indicative skip-grams. To overcome this sparsity problem, we design a transformation for skip-gram features to be more flexible, so that entities can match with skip-grams in a \textit{``softer''} way.

\vspace{-0.3cm}
\begin{table}[htb]
\centering
\caption{A frequency submatrix $\phi$ of president entities and their co-occured skip-grams.}
\label{tab:count_sg}

\scalebox{0.85}{
\begin{tabular}{|c|c|c|c|c|c|}
\hline
Entities &
\makecell{President \\ $\_\_$ and} &
\makecell{President \\ $\_\_$ ,}  &
\makecell{President \\ $\_\_$ said} &
\makecell{President \\ $\_\_$ 's} &
\makecell{President \\ $\_\_$ *}
 \\
\hline
Bill Clinton & 33 & 17 & 2 & 23 & 75\\
Hu Jintao & 9 & 8 & 3 & 0 & 20 \\
Gorbachev & 2 & 3 & 0 & 2 & 7 \\
\hline
\end{tabular}
}
\vspace{-0.3cm}
\end{table}

In Table \ref{tab:count_sg} we list a sub-matrix $\phi$ of $\Phi$ with several president entities and their co-occurred skip-grams. 
% The first four columns of skip-grams are captured by a fixed window size of $(-1, +1)$. In the candidate ranking process, 
The infrequent \textit{``Gorbachev''} does not have enough co-occurrence with type-indicative skip-grams and might be even filtered out from the candidate pool
% ranked lower than expectation. 
What's more, the skip-gram of \textit{``President $\_\_$ said''} might be neglected in the context feature selection process due to its lack of interaction with president entities. 
These four different skip-grams are actually similar since they can all be filled in with a president entity. 
% since their common substring \textit{"President $\_\_$"} does not have much connection with the last term (either \textit{"and"} or \textit{","}) in telling the semantic class of the blank space. 
Therefore, if we can extract the meaningful part from the original window size, such as \textit{``President $\_\_$''} in this example, we will be able to merge these different skip-grams into a common and more flexible pattern: \textit{``President $\_\_$ $\ast$''}, which we named as a \textit{flexgram}. The last term is an asteroid which serves as a wildcard term to be matched with any words. Then we can update the co-occurrence matrix by
\vspace{-0.1cm}
$$\Phi_{e,flex(c)}=\sum_{flex(c')=flex(c)} \Phi_{e,flex(c')}$$
% The co-occurrence of entities and the new flexible skip-gram, which we call \textit{flexgram}, is summed over those fixed skip-grams that allow this kind of transformation.
where flex(c) means the flexible transformation of local context $c$.
Then we can merge those infrequent but type-indicative skip-grams together.
However, one might say a trivial solution to this data sparsity issue is to exhaust a variety of window sizes, (e.g., [-2,+2],[-1,+1],[-1,0],[0,+1]), as has been done by previous studies\cite{Shen2017SetExpanCS,Yu2019CaSE,Rong2016EgoSetEW}. However, this approach very likely ends up generating many skip-grams that are too general. For example, the other half of \textit{``President $\_\_$ and''} is \textit{``$\_\_$ and''}, which only contains stop words, not informative enough in implying the semantic class of the ``skipped'' term. 

Our solution to this problem is to dynamically adjust the window size of a given fixed one, by extracting the independent unit from the the original skip-gram. We denote a skip-gram to be, $l=w_{0}w_{1}...w_{s-1}\_\_ w_{s+1}...w_{n-1}$, where $s$ is the index of the \textit{``skipped''} term. Then we can break down the skip-gram into two parts from $w_i$ and $w_{i+1}$. For example, a skip-gram \textit{``hospital in $\_\_$ has been''} can be broken down in three ways: (1)$l_{left}^1=$\textit{``hospital $\ast$ $\_\_$ $\ast$ $\ast$ ''} and $l_{right}^1=$\textbf{\textit{`` $\ast$ in $\_\_$ has been''}}; (2) $l_{left}^2=$\textbf{\textit{``hospital in $\_\_$ $\ast$ $\ast$ ''}} and $l_{right}^2=$\textbf{\textit{`` $\ast$ $\ast$ $\_\_$ has been''}}; (3)$l_{left}^3=$\textbf{\textit{``hospital in $\_\_$ has $\ast$ ''}} and $l_{right}^3=$\textit{`` $\ast$ $\ast$ $\_\_$ $\ast$ been''}. The bold parts are valid ones for possible flexible transformation that have non-asteroid terms connected with the skipped term.

For each possible transformation from a skip-gram to a \textit{flexgram}, we consider the independence of the left and right part by looking at their pointwise mutual information.

\vspace{-0.2cm}
$$PMI(l_{left}^i,l_{right}^i)=\log \frac{P(l)}{P(l_{left}^i) \cdot P(l_{right}^i)}$$

where $P(l)$ is the occurrence probability of skip-gram $l$ in the corpus.
% in all seen n-grams satisfying that $w_s$ is the skipped term.
% $$P(l) = \frac{Count(w_{0}w_{1}w_{2}\cdots w_{s-1}\_\_w_{s+1} \cdots w_{n-1})}{\sum_{w_{0},w_{1},w_{2},\cdots,w_{s-1},w_{s+1},\cdots,   w_{n-1}}Count(w_{0}w_{1}w_{2}\cdots w_{s-1}\_\_w_{s+1} \cdots  w_{n-1})}$$
% $$P(l) = \frac{Count(w_{0}w_{1}w_{2}\cdots w_{s-1}\_\_w_{s+1} \cdots w_{n-1})}{\sum_{|l'|=n,w_s="\_\_"} Count(l')}$$
% $P(l_{left}^i)$ and $P(l_{right}^i)$ can be derived in the same way. 
% We first select the broken pairs with the lowest PMI score, and then set a threshold $\gamma$ so that when
If the most independent pair satisfies
$$\min_i PMI(l_{left}^i,l_{right}^i)<\gamma,$$ 
we break the original skip-gram $l$ into $l_{left}^i,l_{right}^i$. In this example, suppose we have $l_{left}^2=$\textbf{\textit{``hospital in $\_\_$ $\ast$ $\ast$ ''}} and $l_{right}^2=$\textbf{\textit{`` $\ast$ $\ast$ $\_\_$ has been''}} to be the feasible breaking pair, we then look into the valid parts in $l_{left}^i$ and $l_{right}^i$, and compare their occurrences with the original skip-gram. If
% $$\frac{Count(l_{left}^i)}{Count(l)}>k,$$ 
$$\frac{\sum_{e'} \Phi_{e',l_{right}^i}}{\sum_{e'} \Phi_{e',l}} > k$$
then we can say $l_{right}^i$ generalizes too much from the original skip-gram $l$ and prevent the transformation (as is the skip-gram ``\textbf{\textit{$\ast$ $\ast$ $\_\_$ has been}}'' in this example), otherwise, we will allow the flexible transformation. 
% In our experiment we set $k=100$. In our running example, $Count(\textit{"hospital in \_\_ had been "})=1$, $Count(\textit{"hospital in \_\_ * * "})=70$ and $Count(" * * \_\_ has been")=2053$, consider that $l_{right}^i$ generalize too much from the original skip-gram and might not capture the type-indicative part, we transform the original skip-gram $l=\textit{"hospital in \_\_ had been "}$ into $l_{right}^i=\textit{"hospital in \_\_ * * "})$ which still infers that the skipped term is a place that holds a hospital.

We perform this step of transformation for skip-gram contexts in the pre-processing part before the iterations of entity expansion, thus only requiring one-time calculation and will not harm the time efficiency.

\vspace{-0.1cm}
\subsubsection{Combining Different Context Features by Rank Ensemble}

% and distributed representations such as Word2Vec. The features are complimentary in that skip-grams require "hard match" between candidate and pattern so the matched entities have a high precision but low coverage of the whole set. Distributed representations that learn a low-dimensional vector space to capture semantics of terms, however, capture "softer" similarity between entities since they do not impose hard positional constraints in the local context window. Though more flexible, only using term embedding would result in generating related but not parallel concepts, such as entities belonging to parent or child classes. 

We use both skip-gram based features and distributed representation to measure the similarity between candidates and seed entities. These two features complement each other in that skip-gram based features require exact matching of textual patterns and impose stronger positional constraint, leading to fewer entities being matched, and embedding based features tend to capture more semantically similar terms while allowing more noise. Therefore, combining these two features improves the final result, as is shown in several previous studies \cite{Shen2017SetExpanCS,Yu2019CaSE}. For skip-gram based features, we first extract a fixed window size of  $[-3,+3]$ around each entity mention in the corpus, and then transform the skip-grams dynamically into shorter flexgrams as is proposed in Section \ref{sec:flexgram}. After the transformation, co-occurrence between entity mentions and context features are recalculated by summing over same flexgrams. For embedding based features, we incorporate BERT, a recently introduced pre-trained language model for contextualized word representation \cite{Devlin2019BERTPO}, which has shown to be very effective on several NLP tasks including set expansion \cite{Yu2019CaSE}. To get a context-free embedding for each term, we use the pre-trained model to first get the contextualized embedding of an entity mention at each occurrence, and then average over all the occurrences in the corpus to get its center point in the embedding space.

We separately score the candidates based on skip-gram and embedding features. 
\vspace{-0.1cm}
$$score_{sg, S}(e)=\frac{1}{|S|} \sum_{e'\in S} Sim(e, e'|F^*)$$
\vspace{-0.1cm}
$$score_{emb, S}(e)=\frac{1}{|S|} \sum_{e'\in S} cos(v_e, v_{e'})$$
skip-gram features based similarity is defined in Equation \eqref{eq:sim}, while embedding based similarity is calculated by cosine similarity. We remove candidates if $score_{S}(e)$ is smaller than any $score_{S_k}(e)$ where $S_k \in C_{aux}$. We then create two pre-ranking lists of candidates based on their score from both skip-gram and embedding features, and use rank ensemble \cite{Shen2017SetExpanCS} to combine results from both rankings to get the final score for each candidates. 
\vspace{-0.1cm}
$$MRR(e)=\frac{1}{r_{sg}(e)}+\frac{1}{r_{emb}(e)}$$
Mean reciprocal ranking result based on the pre-ranking lists of two separate features, is used for sorting. The expansion algorithm naturally stops when all candidates have a lower similarity with seeds in target set than that with seeds in auxiliary set(s). Algorithm \ref{alg:mult_expan} describes the Multiple Sets Co-Expansion module.

\small
\vspace{-0.3cm}
\begin{algorithm}
    \SetKwInOut{Input}{Input}
    \SetKwInOut{Output}{Output}

    \Input{Initial seed sets $S_1, S_2, ..., S_N$
    \newline Entity set $E$; 
    \newline Context feature set $F$; 
    \newline Max number of expansion iterations $T$; 
    \newline Number of entities generated in one iteration $t$;
    \newline Number of context features used in one iteration $Q$ }
    \Output{The expanded sets $X_1, X_2, ..., X_N$}
    
    \For{ $i$ in 1..$N$ }
    {
    	$X_i \gets S_i$
    }
      
    \For{ $iter$ in 1..$T$ }
    {
    	$E \gets E - X$
    	
    % 	$F_{ranked} \gets$ ranked list of $f \in F$ 
    	
    	$F^* \gets $ top $Q$ items based on Equation \eqref{eq:cf_sel}
    	
    	\For{ $i$ in 1..$N$ }
    	{
    		$score_{sg, X_i}(e) \gets \sum_{x \in X_i} \frac{1}{|X_i|} Sim(e, x | F^*) $
    		
    		$r_{sg, X_i}(e) \gets$ the rank of $e \in E$ based on $score_{sg, X_i}(e)$
    		
    		$score_{emb, X_i}(e) \gets \sum_{x \in X_i} \frac{1}{|X_i|} cos(v_e, v_x) $
    	
    		$r_{emb, X_i}(e) \gets$ the rank of $e \in E$ based on $score_{emb, X_i}(e)$
    		
    		$score(e) \gets \frac{1}{r_{sg, X_i}(e)} + \frac{1}{r_{emb, X_i}(e)}$
    	
    		$E_{ranked, X_i} \gets$ ranking list of $e \in E$ based on $score(e)$
    	
    		$X_i \gets X_i \cup $ the top $t$ items in $E_{ranked, X_i}$
    	}

    }

    Return $X_1, X_2, ..., X_N$ 
     
	\caption{Multiple Sets Co-Expansion}
	\label{alg:mult_expan}
\end{algorithm}
\normalsize

% \vspace{-0.3cm}
\section{Experiments and Results}
\subsection{Experiment Setup}

% \begin{table}[ht]
% \caption{MAP@50 of Full Model vs. Full Model (no aux.) on Different Semantic Classes in \textit{Wiki} Dataset.}
% %\vspace{-0.3cm}
% \label{tab:breakdown}
% \scalebox{1.0}{
% \begin{tabular}{|c|c|c|}
% \hline
% Semantic Class & Full Model  & Full Model (no aux.)\\
% \hline
% Countries & 0.990 & 0.968\\
% \hline
% TV Channels & 0.901 & 0.847\\
% \hline
% US States & 0.935 & 0.930 \\
% \hline
% Diseases & 0.979 & 0.963\\
% \hline
% Sports Leagues & 0.879 & 0.860\\
% \hline
% Chinese Provinces & 0.702 & 0.680 \\
% \hline
% Political Parties & 0.982 & 0.927 \\
% \hline
% Companies & 0.876 & 0.709 \\
% \hline
% \end{tabular}
% }
% %\vspace{-0.3cm}
% \end{table}

\begin{table}[htb]
    \centering
    \scalebox{0.9}{
    \begin{tabular}{|c|c|c|c|c|}
    \hline
    Dataset & $\#$ classes & $\#$ queries & entity vocabulary size & $\#$ documents \\
    \hline
    Wiki & 8 & 40 & 41242 & 780556 
    % & \makecell{Countries, TV Channels, Political Parties, Sports Leagues,\\ Chinese Provinces, Diseases, US States, Companies}
    \\
    \hline
    APR & 3 & 15 & 71707 & 1014140 
    % & Countries, Political Parties, US States
    \\
    \hline
    \end{tabular}}
    \caption{Descriptions of datasets.}
    \vspace{-0.3cm}
    \label{tab:data}
\end{table}

\subsubsection{Datasets}
We conduct our experiments on two large corpus also used in previous studies \cite{Shen2017SetExpanCS}: (1) \textbf{Wiki} is a subset of Wikipedia English dump from May 2011. (2) \textbf{APR} contains news articles published by two news platforms, Associated Press and Reuters in the year of 2015. We use the same sets of seed queries and ground truth of each semantic classes with previous studies as well. For \textbf{Wiki} dataset, there are 8 semantic classes, and 5 queries per class. For \textbf{APR} dataset, there are 3 semantic classes, and 5 queries for each class. These queries cover a wide range of semantic classes. Each query contains three seed terms belonging to a certain semantic class. Table \ref{tab:data} lists the detailed descriptions of these datasets and queries\footnote{The code and data are available at https://github.com/teapot123/SetCoExpan}.

To extract possible entity mentions, we use the same preprocessing pipeline as SetExpan \cite{Shen2017SetExpanCS}. 
% We learn the distributed vector space for related terms retrieval in Section \ref{sec:emb} beforehand, so there is no need to retrain the embedding space in each iteration.

\subsubsection{Parameter Settings}
Since our framework iteratively generates auxiliary sets and co-expands target and auxiliary sets, we set the expansion number $t$ at each iteration to be 5, and will later provide parameter study of $t$. In the auxiliary sets generation module, the embedding training balances the global context loss and local context loss by a ratio $\lambda$ of 1.5, and we retrieve 10 related words for each element in the current expanded set.  In the cross-seed merging stage, two groups can be merged together if they have enough overlap, thus we set $\eta$ to be
$\sqrt{min(g^i_{e,expanded}, g^j_{e,expanded})}$.
In the multiple sets co-expansion module, since the size of generated auxiliary sets are often larger than the currently expanded target set, we need to balance the size of multiple sets to prevent the model from extracting features that only focus on making the largest set to be cohesive. We randomly sample seeds from each auxiliary set to keep it as large as the target set.
When extracting skip-gram features for each entity mention, we first extract a fixed window size of 3, and then transform them into dynamic independent units as described in section \ref{sec:flexgram}, and there are two parameters in this transformation: the breaking threshold $\gamma$ is set to be 1.0, and the generalizing threshold $k$ is set to be 100, which stops the original skip-gram to be broken down to too general contexts. In the Algorithm \ref{alg:mult_expan}, we set $T=10$ and $Q=200$ as SetExpan \cite{Shen2017SetExpanCS}. We use the same hyperparameters for all queries on both datasets.

\subsubsection{Evaluation Metric}
We use Mean Average Precision (MAP) to evaluate the ranking result of our methods as well as other baseline methods. For a query $q$ with 3 seeds, we take the first k output entities from each algorithm as list $L$ and calculate Average Precision at $k=10,20,50$, defined as
{\small
$$AP_k(L)=\sum_{i=1}^k P(k)\Delta r(k)$$
}
where $P(k)$ denotes precision$@k$ and $\Delta r(k)$ denotes the change in recall from the $(k-1)^{th}$ term to the $k^{th}$ term. Therefore, only the terms expanded correctly will contribute to this metric.
Mean Average Precision at $k$ is defined as the mean of Average Precision at $k$ over all queries in all semantic classes.
{\small
$$MAP_k = \frac{\sum_L AP_k(L)}{|L|} $$
}

\begin{table*}
\centering
\caption{Mean Average Precision across all queries on \textit{Wiki} and \textit{APR}.}
\label{tab:map}
\begin{tabular}{c|ccc|ccc}
\toprule
\multirow{2}{*}{Methods} &
\multicolumn{3}{c|}{\textit{Wiki}} &\multicolumn{3}{c}{\textit{APR}} \\
% \cline{2-7}
& MAP@10 & MAP@20 & MAP@50
& MAP@10 & MAP@20 & MAP@50\\
\midrule
CaSE & 0.897 & 0.806 & 0.588 & 0.619 & 0.494 & 0.330 \\
SetExpander & 0.499 & 0.439 & 0.321 & 0.287 & 0.208 & 0.120 \\
SetExpan & 0.944 & 0.921 & 0.720 & 0.789 & 0.763 & 0.639  \\
BERT & 0.970 & 0.945 & 0.853 & 0.890 & 0.896 & 0.777 \\
\midrule
\CoExpan (no aux.) & 0.964 & 0.950  & 0.861 & 0.900 & 0.893 & 0.793 \\
\CoExpan (no flex.) & 0.973 & 0.961 & 0.886 & 0.927 & 0.908 & 0.823 \\
\CoExpan & \textbf{0.976} & \textbf{0.964} & \textbf{0.905} & \textbf{0.933} & \textbf{0.915} & \textbf{0.830} \\
\bottomrule
\end{tabular}
\vspace{-0.3cm}
\end{table*}

\vspace{-0.2cm}
\begin{figure*}[htb]
    \centering % <-- added
\begin{subfigure}{0.32\textwidth}
  \includegraphics[width=\linewidth]{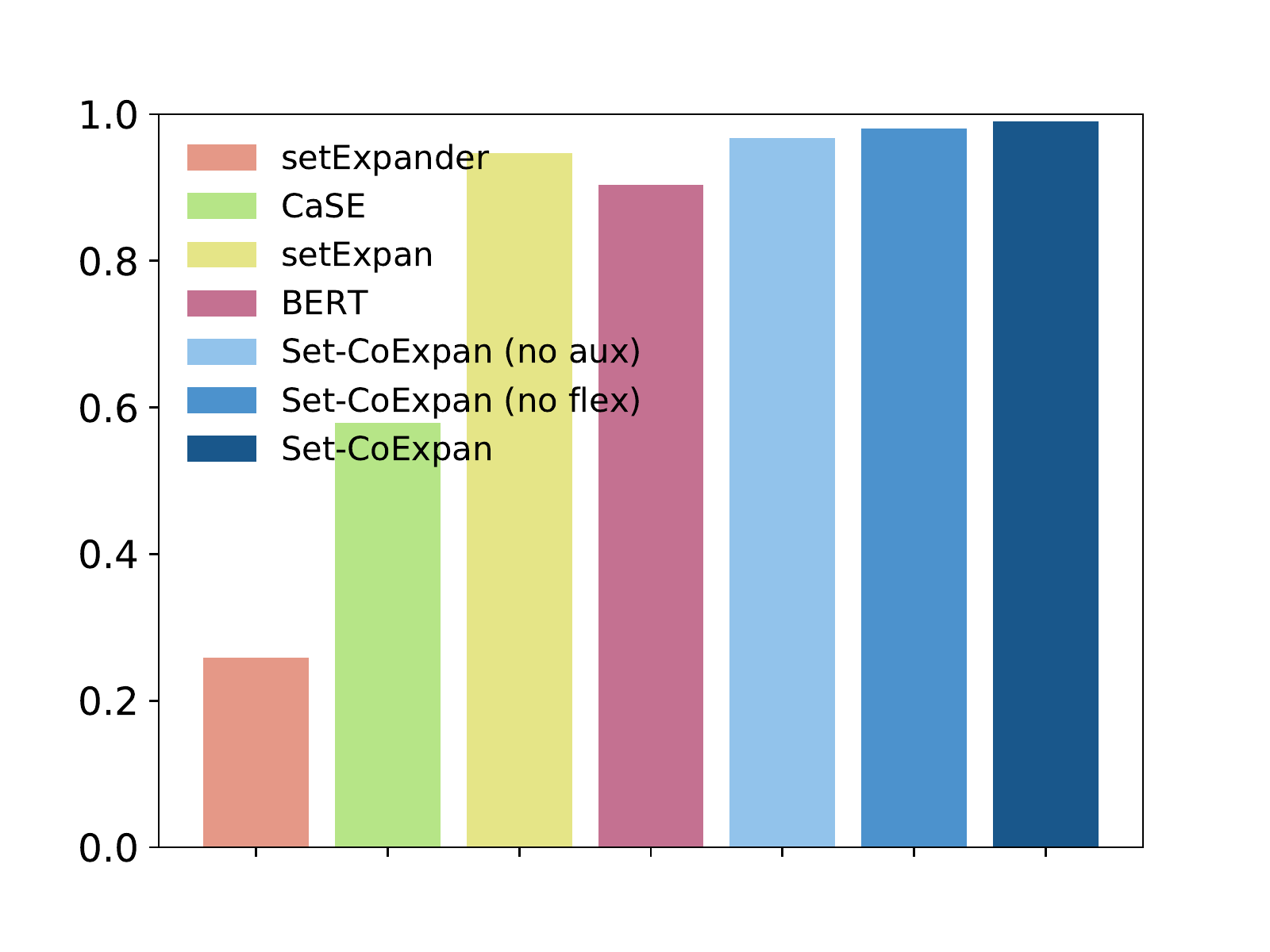}
  \caption{Wiki: countries}
  \label{fig:1}
\end{subfigure}\hfil % <-- added
\begin{subfigure}{0.32\textwidth}
  \includegraphics[width=\linewidth]{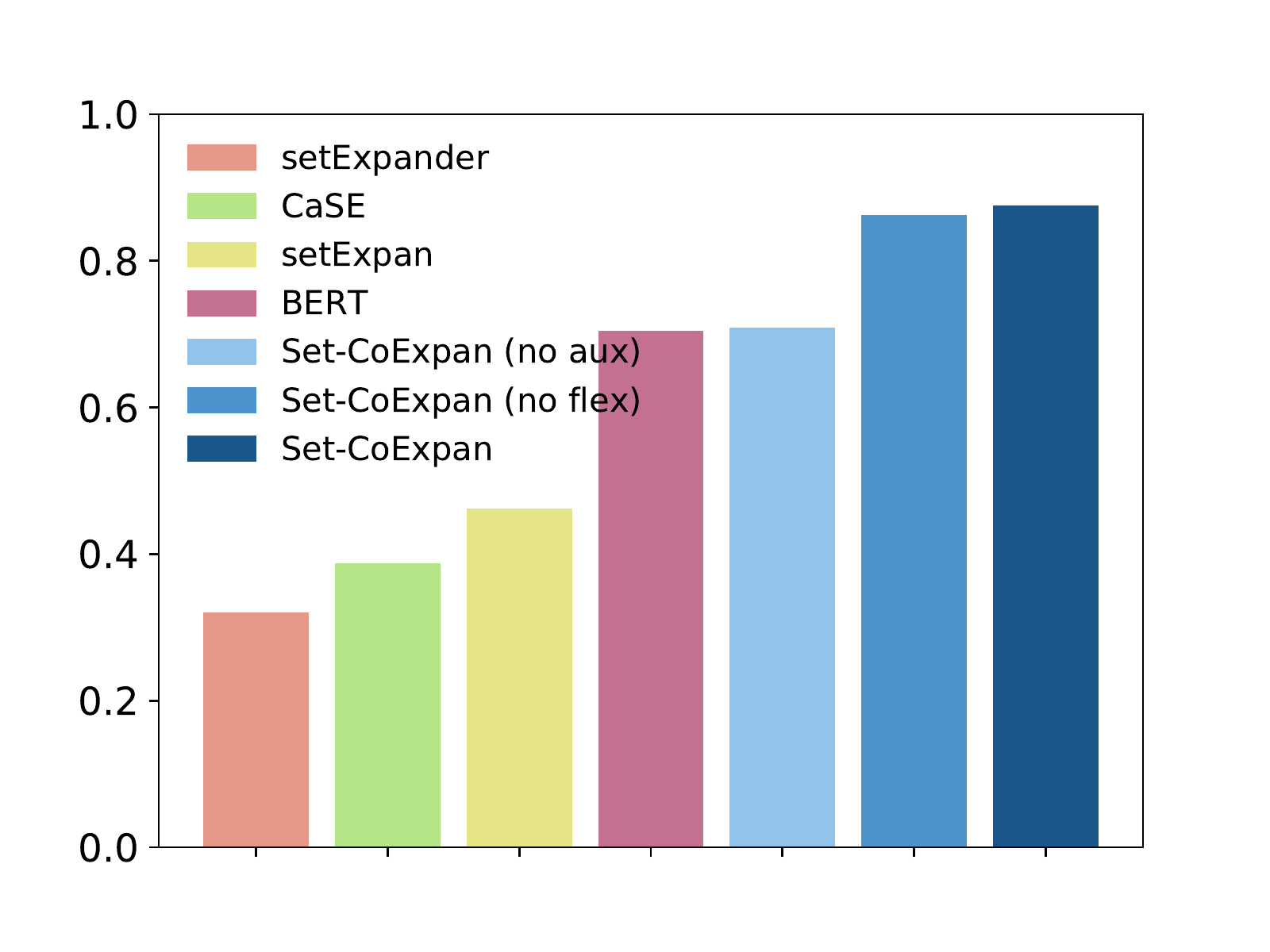}
  \caption{Wiki: companies}
  \label{fig:2}
\end{subfigure}\hfil % <-- added
\begin{subfigure}{0.32\textwidth}
  \includegraphics[width=\linewidth]{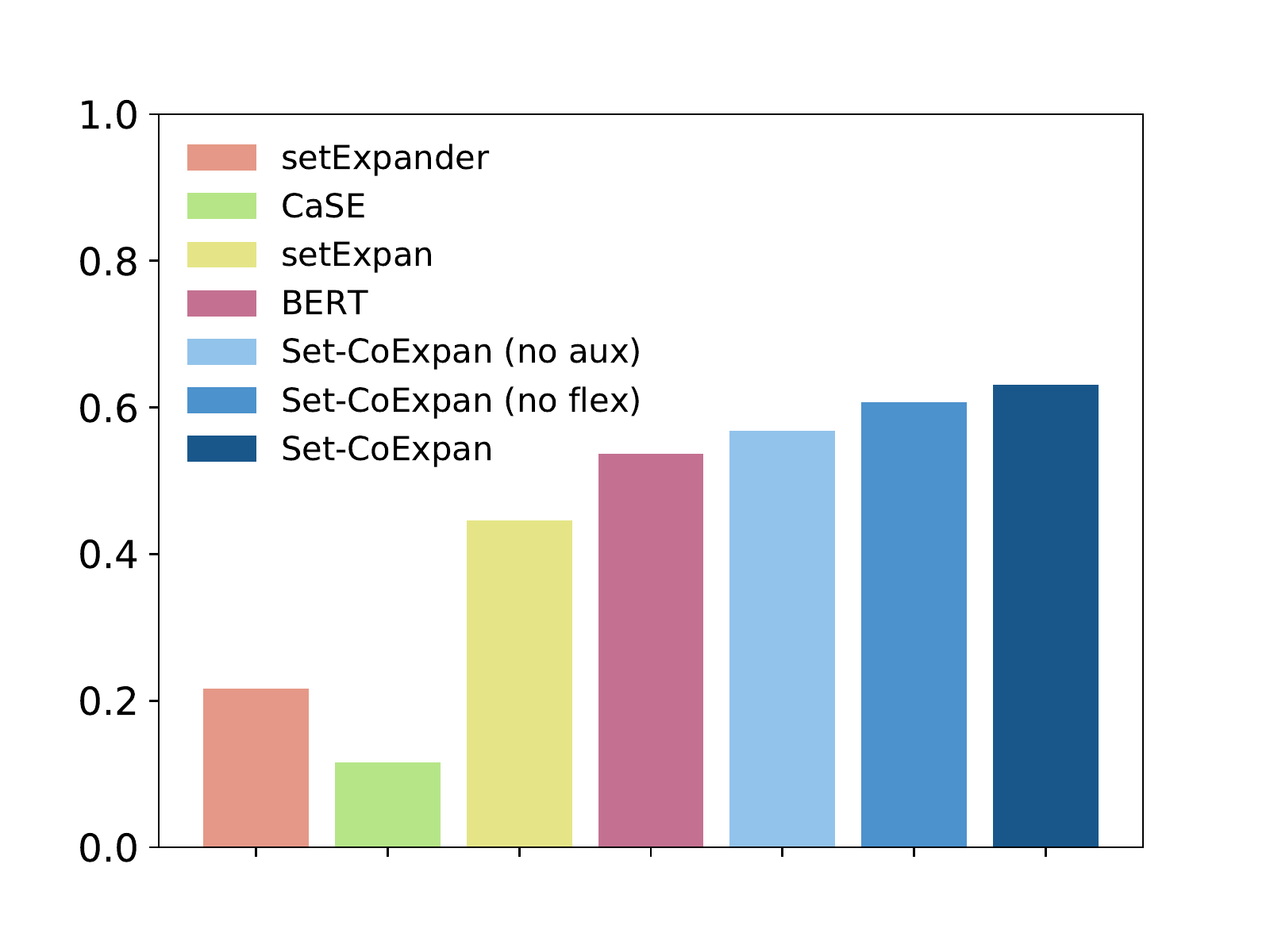}
  \caption{APR: Political Parties}
  \label{fig:3}
\end{subfigure}

\medskip
\begin{subfigure}{0.32\textwidth}
  \includegraphics[width=\linewidth]{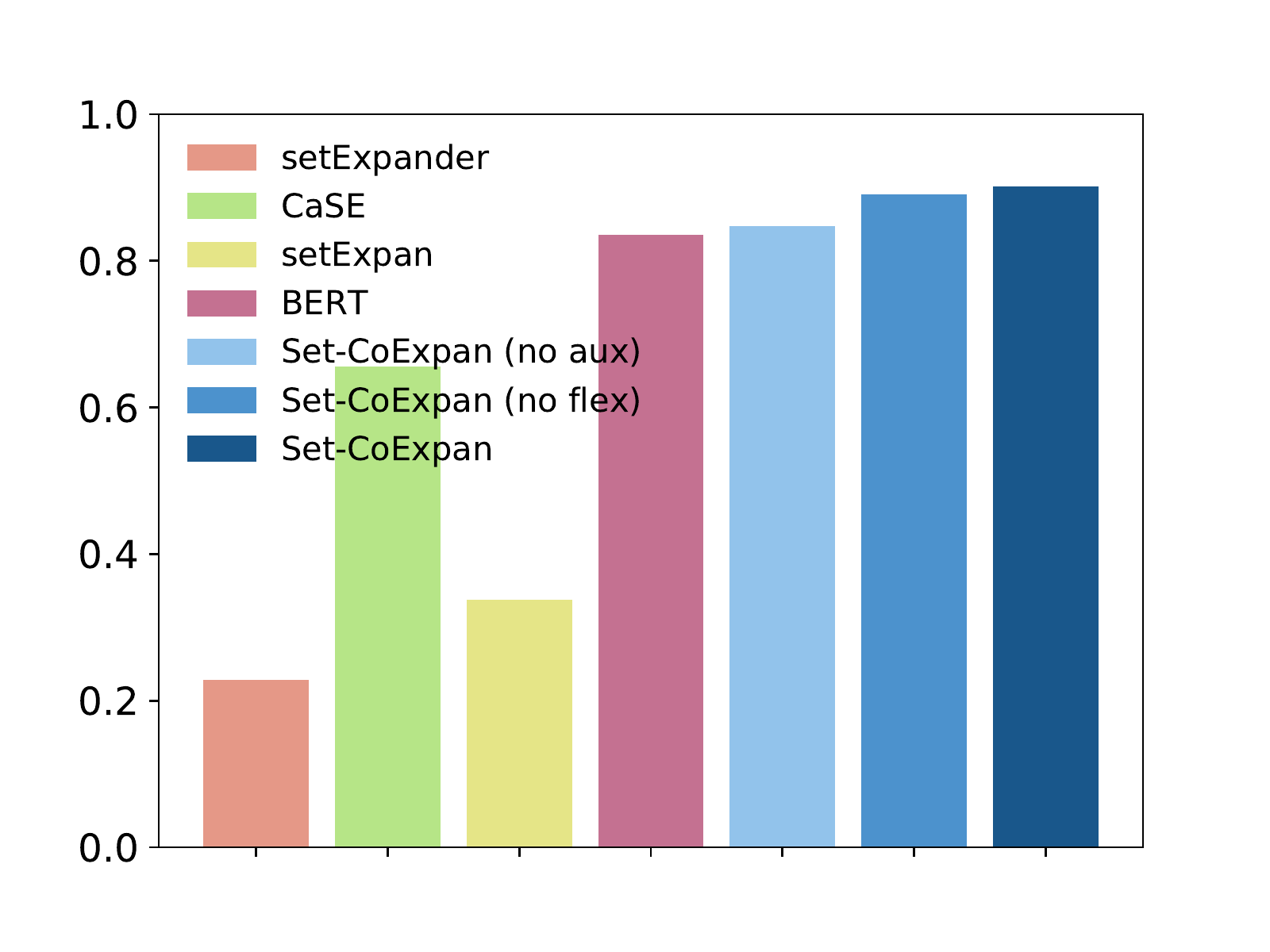}
  \caption{Wiki: TV channels}
  \label{fig:4}
\end{subfigure}\hfil % <-- added
\begin{subfigure}{0.32\textwidth}
  \includegraphics[width=\linewidth]{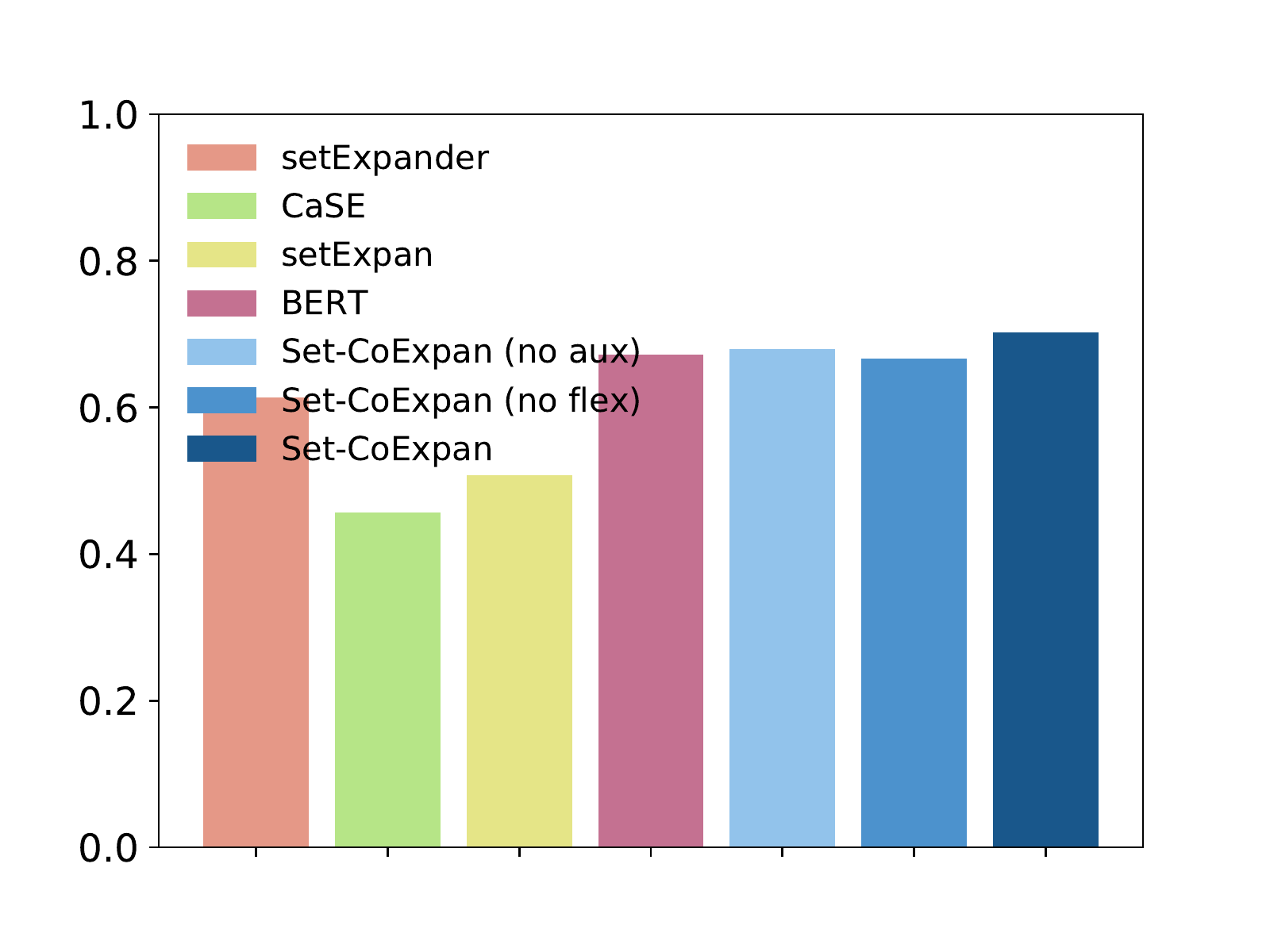}
  \caption{Wiki: china provinces}
  \label{fig:5}
\end{subfigure}\hfil % <-- added
\begin{subfigure}{0.32\textwidth}
  \includegraphics[width=\linewidth]{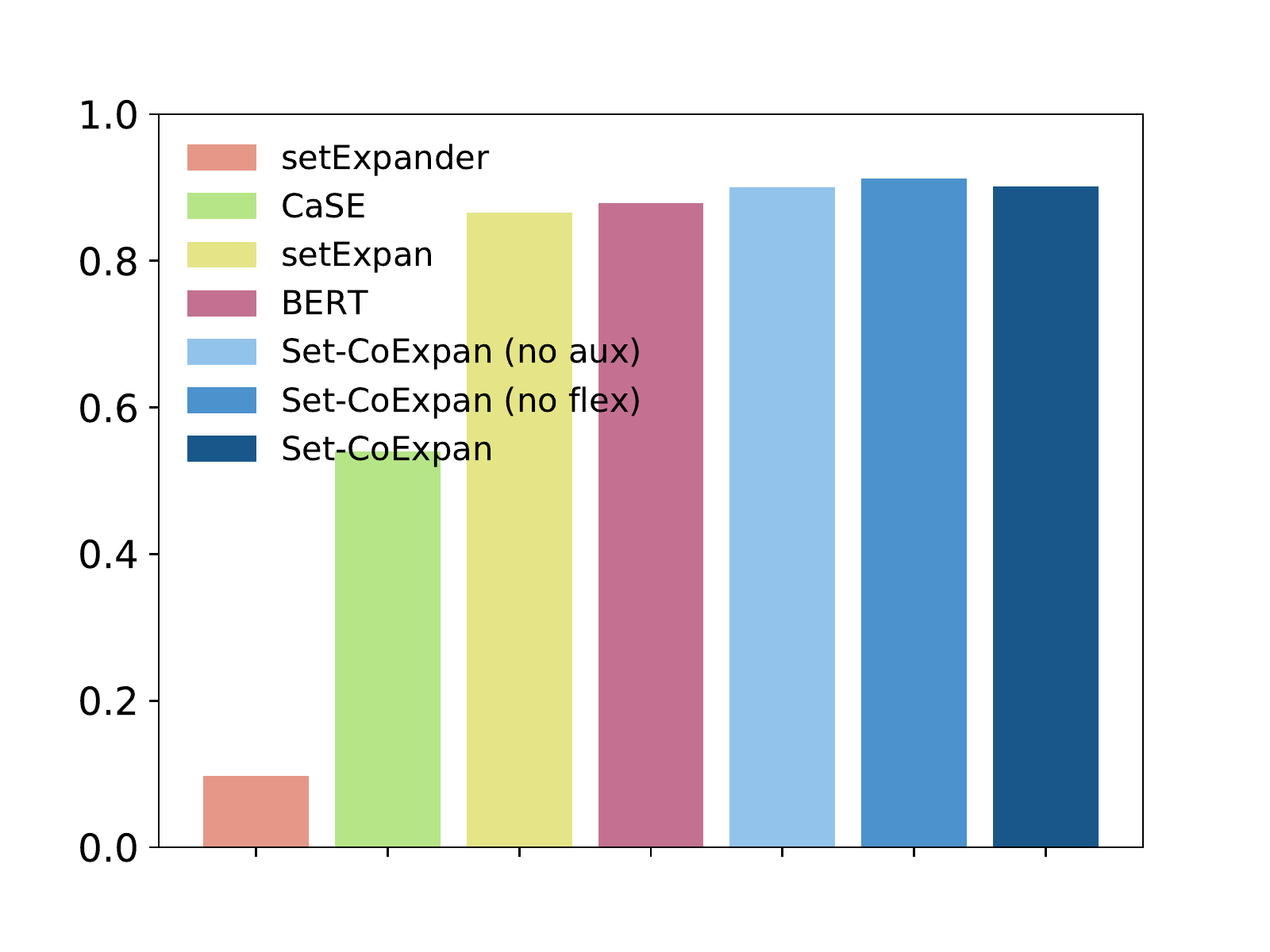}
  \caption{APR: US states}
  \label{fig:6}
\end{subfigure}
\caption{Performance comparison on different semantic classes in \textit{Wiki} and \textit{APR}.}
\vspace{-0.3cm}
\label{fig:breakdown}
\end{figure*}

\vspace{-0.3cm}
\subsection{Compared Methods}
We compare our methods to several previous corpus-based set expansion algorithms that only take corpus as input to expand entity seed sets. We also implement a baseline named BERT that only uses the pre-trained BERT model to retrieve the most similar entities. To show the effectiveness of auxiliary sets and flexible transformation of skip-grams in improving set expansion, we also compared \CoExpan with several ablations.

\begin{itemize}[leftmargin=*]
% 	\item EgoSet \cite{Rong2016EgoSetEW}: This method is originally designed for multi-faceted set expansion tasks. It first constructs a ego-network for seed entities, and then uses a community detection module to obtain different senses. To align with our setting, we omit the community detection module and treat each entity as a single sense term.
	\item SetExpan \cite{Shen2017SetExpanCS}: This method uses skip-gram features, distributed representations and coarse-grained types as context features. It uses rank ensemble to combine the result of three pre-ranking lists. We run the algorithm using its default setting.
	\item SetExpander \cite{Mamou2018setExpander}: This method incorporates multiple context features by training separate embedding space for each type of contexts, and a classifier is trained on a labeled dataset to adjust the weight of different features. 
% 	We use its pre-trained model on our corpus.
	\item CaSE \cite{Yu2019CaSE}: This method combines skip-gram patterns and distributed representations to generate a one-time ranking for candidate entities. The model has three variants using different distributed representations, and we report the result of CaSE-W2V which has the highest performance.
	\item BERT \cite{Devlin2019BERTPO}: We implement a baseline that only uses BERT, a recent contextualized embedding framework. We use the pre-trained model (uncased, base, 768 dimensions) and average over all occurrences in the corpus to get the centroid for any entity. We then use KNN to search for most similar entities.
	\item \CoExpan (no aux.): This ablation excludes the auxiliary sets generation module thus only expand one query of seed set in the co-expansion module.
	\item \CoExpan (no flex.): This ablation expands auxiliary sets simultaneously with the target set. However, it removes the flexible transformation of skip-grams.
\end{itemize}

\vspace{-0.1cm}
\begin{table*}
\centering
\caption{Auxiliary sets generated for various queries.}
\label{tab:aux}
\begin{tabular}{|c|c|c|}
\hline
Class & Query & Auxiliary sets\\
\hline
Companies & Myspace, Youtube, Twitter & \makecell{ 
% 	\textbf{Core (Companies):} Facebook, Ebay, LiveJournal, Amazon, Google\\ 
	    \textbf{(Products):} flickr, wordpress, google earth, gmail, google maps}\\
\hline
Countries & Australia, France, Germany & \makecell{
% \textbf{Core (Countries):} Italy, Canada, Belgium, Australia, Spain\\ 
	\textbf{(Provinces):} Queensland, New South Wales, Saxony, Bavaria, Thuringia \\
	\textbf{(Cities):} Brisbane, Canberra, Rennes, Hamburg, Stuttgart} \\

\hline
TV Channels & ESPN News, ESPN Classic, ABC & \makecell{
% \textbf{Core (TV Channels):} STAR Plus, FOX SPorts Net, Comedy Central, CBS, NBC\\
	\textbf{(TV Programmes):} the young and the restless, \\all my children, guiding light, general hospitale}\\

\hline
Sports Leagues & \makecell{national football league, \\ national hockey league, \\ major league baseball} & \makecell{
% \textbf{Core (Sports Leagues):} american football league, national basketball association, \\
% canadian football league, arena football league, international hockey league
%     \\
	\textbf{(Sports Teams):} new york jets, ottawa senators,\\ chicago white sox, dallas cowboy, st.louis hawks}\\
\hline	
Political Parties & \makecell{new democratic party, \\ liberal party of canada, \\ northern ireland labour party} & \makecell{
% \textbf{Core (Political Parties):} Canadian Alliance, Australian Labor Party, \\Ulster Unionist Party, Irish Parliamentary Party, Conservative Party\\
	\textbf{(Elections):} 1980 federal election, 1997 federal election, 1980 election, \\1962 election, 2008 provincial election}\\
\hline
Chinese Provinces & \makecell{jiangsu, liaoning, sichuan} & \makecell{
% \textbf{Core (China Provinces):} henan, zhejiang, yunnan, shandong, jiangxi\\
	\textbf{(China Cities):} xi'an, hangzhou, shanghai, chengdu, beijing}\\
\hline
Diseases & \makecell{tuberculosis, \\ parkinson's disease, \\ esophageal cancer} & \makecell{
% \textbf{Core (Diseases):} lung cancer, leukemia, breast cancer, alzheimer's disease, aids\\
	\textbf{(Symptoms):} tumor, dehydration, dementia, muscle stiffness}\\
\hline
US States & \makecell{Texas, Florida, New Mexico} & \makecell{
% \textbf{Core (US States):} California, Alabama, Virginia, Iowa, Maryland\\
	\textbf{(US Cities):} fort worth, san antonio, jacksonville, tampa, orlando}\\
\hline
\end{tabular}
\end{table*}

\vspace{-0.2cm}
\subsection{Results and Discussions}

\subsubsection{Overall Performance on Two Datasets}
For our method and all baseline methods, we use the same queries in the datasets, and report $MAP@k$ ($k=10,20,50$) in Table \ref{tab:map}. The result clearly shows that our model has the best performance over all the methods in both datasets. Among all the baseline methods, BERT is the strongest one,  since it stores a very large pre-trained language model to represent word semantics in a piece of context, which also explains why we have a very large margin over previous iterative-based methods like SetExpan when we incorporate BERT as embedding-based features. However, BERT itself could not outperform our model, which implies that modeling distributional similarity alone is not enough for generating accurate expansion. 
% We also observe that SetExpan is the strongest among the baselines and SetExpander has a rather low performance because its classifier is a pre-trained one, and might not adapt well to other corpus. CaSE provides a one-time ranking result, which sacrifices performance in pursuit of efficiency. BERT has a rather stable performance over different length of ranking lists, but it is still not better than our method, which shows that distributional 
We can also observe that \CoExpan also outperforms the ablations without auxiliary sets or flexgram,
which justifies that auxiliary sets indeed improve the set expansion process by providing clearer discriminative context features for multiple sets to prevent them from expanding into each other's semantic territory. The advantage of \CoExpan over the ablation without flexgram also proves that it is necessary to cope with the data sparsity problem by extracting independent units from a fixed local context window. In this way, we relieve the hard-matching constraint brought by skip-gram features by making it more flexible. What's more, \CoExpan has a larger advantage over all the baselines when the ranking list is longer, which indicates that when the seed set gradually grows out of control and more noises appear, our method is able to steer the direction of expansion and to set barriers for out-of-category words to come in.

\begin{figure}[tb]
    \centering % <-- added
\begin{subfigure}{0.49\linewidth}
  \includegraphics[width=\linewidth]{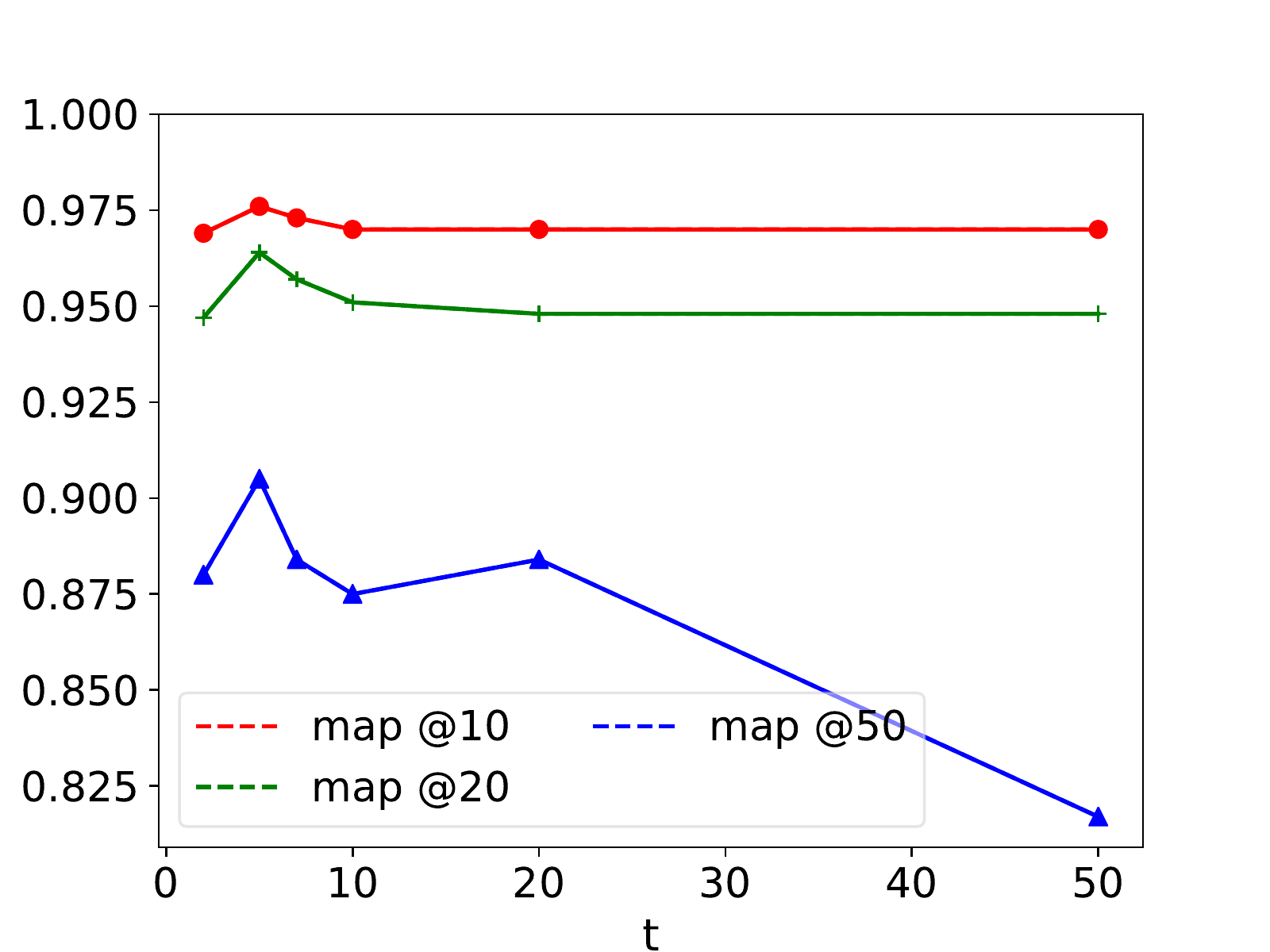}
  \caption{Wiki}
  \label{fig:1}
\end{subfigure}\hfil % <-- added
\begin{subfigure}{0.49\linewidth}
  \includegraphics[width=\linewidth]{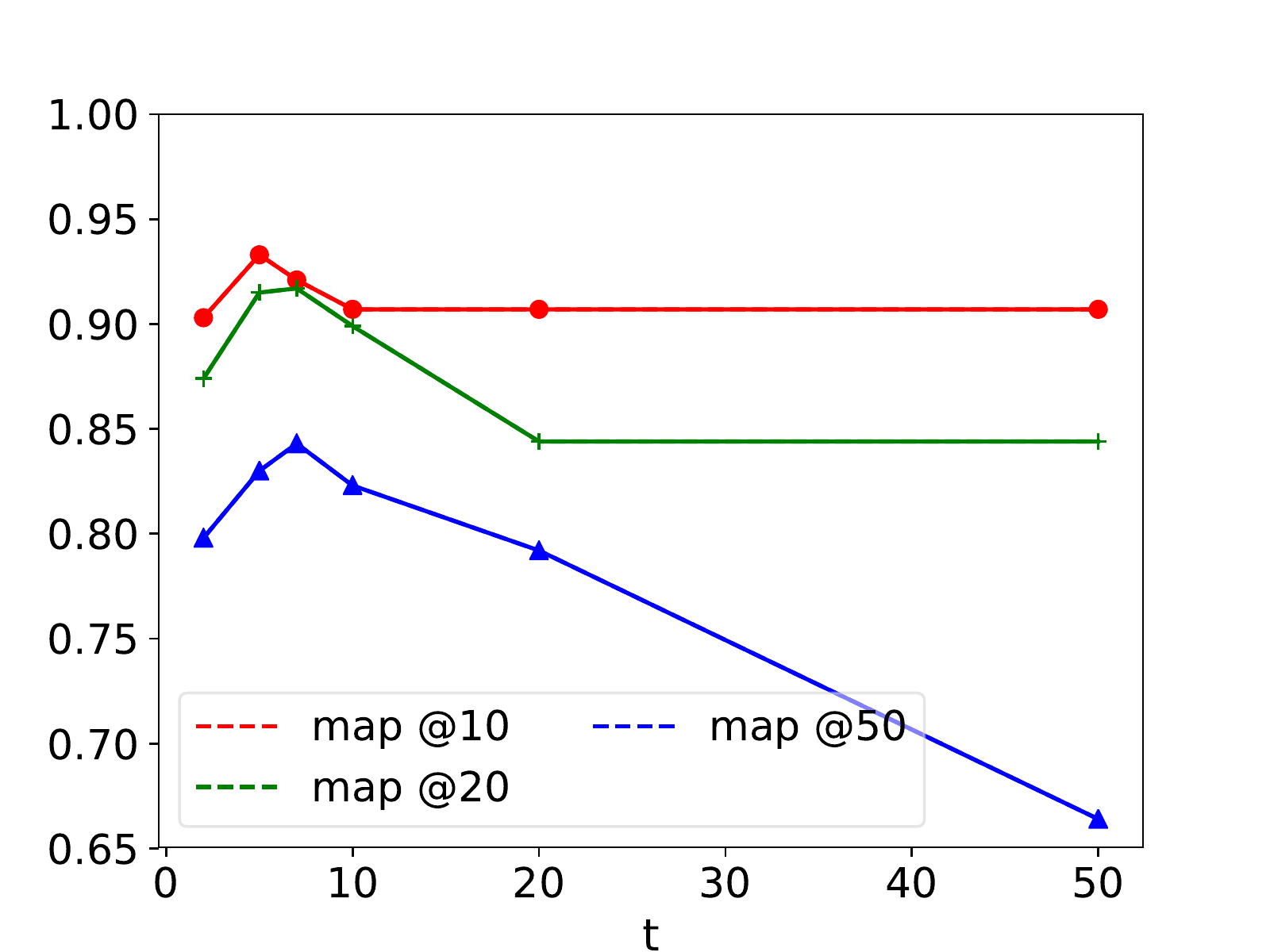}
  \caption{APR}
  \label{fig:3}
\end{subfigure}

\caption{Parameter Study.}
\label{fig:parameter}
\vspace{-0.3cm}
\end{figure}

\vspace{-0.2cm}
\subsubsection{Performance Comparison on Different Semantic Classes}
% We further break down the results on \textit{Wiki} Dataset into different semantic classes to demonstrate the power of adding auxiliary sets. In Table \ref{tab:breakdown}, we show the comparison of our full model and the ablation without auxiliary sets in each class. In the semantic class level, we can see for classes like "\textit{Countries}", "\textit{TV Channels}" and "\textit{Companies}", the auxiliary sets help improve the performance to a great extent. On the other hand, for some categories like "\textit{US States}", "\textit{Diseases}" and "\textit{Sports Leagues}", the improvement is marginal, which implies that in these situations our method does not generate meaningful auxiliary sets to learn from. Since our method of generating auxiliary sets is entirely unsupervised, it is not guaranteed that each semantic class can find some related satellite sets.

We further break down the results on different semantic classes to see which method achieves the best performance on each semantic class, as shown in Figure \ref{fig:breakdown}. We can see that the performances of several baseline methods can be rather unstable over different semantic classes, while our method can outperform them in all cases. For example, setExpander gives good results in \textit{China Provinces} but bad results for other classes. This is because the weights in its classifier is pre-trained on other corpus with supervision, and might not adapt well to copora without labels for training. setExpan shows promising results in \textit{Countries} and \textit{US States}, but still suffers from the problem of semantic drift in other categories.

% One reason is that all baseline methods are lack of negative supervision. They may wrongly output some entities from closely related but not the same semantic classes since they may share some similar context features. In comparison, our method can capture some of these entities in the auxiliary sets, and thus avoid this kind of mistake. Another reason is that many of the baseline methods make use of skip-gram features. This may introduce data sparsity problem because Two entities can share a skip-gram feature only if their appear in exactly same context. In our method, we refine the skip-gram feature by making it more flexible. This makes the co-occurrence of context features easier, and thus mitigate the data sparsity problem.

The margin between \CoExpan and the two ablations also varies between semantic classes. We can observe that for classes like ``\textit{Countries}'', ``\textit{TV Channels}'', ``\textit{Political Parties}'' and ``\textit{Companies}'', the auxiliary sets help improve the performance to a great extent. On the other hand, for some categories like ``\textit{China Provinces}'' and ``\textit{US States}'', the improvement is marginal, which implies that in these situations our method does not generate meaningful auxiliary sets to for the target set to compare with. Since our method of generating auxiliary sets is entirely unsupervised, it is not guaranteed that each semantic class can find very meaningful auxiliary sets.

\begin{table*}
\centering
\caption{Results of Co-Expansion and Separate Expansion of Target Set and Auxiliary Sets.}
\label{satellite}

\scalebox{0.95}{
\begin{tabular}{| c | c c | c c | c c |}
 
%   \hline 
%   Target Class  & \multicolumn{6}{c|}{Countries} \\
  \hline 
  \makecell{seeds} &  \multicolumn{2}{c|}{\makecell{
            seeds from Target Set:
            \\Australia, France, Germany 
            }}
             & \multicolumn{2}{c|}{\makecell{
            seeds from Aux. Set 1:
            \\Queensland, Saxony, New South Wales
            }}
             & \multicolumn{2}{c|}{\makecell{
            seeds from Aux. Set 2:
            \\Brisbane, Canberra, Stuttgart
            }}
  		    
  		    \\

  \hline
  \multirow{5}*{\makecell{Multiple Sets\\ Co-Expansion}}      
  		% & Italy & Hesse   & Berlin \\
        % & Luxembourg & Baden-Wurttemberg  & Hanover \\
       	% & Canada & Baden  & Dortmund \\
       	% & Belgium & Saxony-Anhalt     & Frankfurt \\
       	% & Norway & Schleswig-Holstein   & Heidelberg \\
        % & Spain & Rhineland-Palatinate     & Strasbourg  \\
       	% & The Netherlands & Silesia (\xmark)     & Munich	\\
       	% & Denmark & WestPhalia  & Bonn \\
       	% & England & Mecklenburg-Vorpommern     & Cologne \\
       	% & Switzerland & Saarland     & Mannheim \\
       	
       	& Italy & Luxembourg 
       	& Baden-Wurttemberg & Hesse  
       	& Berlin & Hanover \\
       	
       	& Canada & Belgium 
       	& Baden & Saxony-Anhalt 
       	& Dortmund & Frankfurt \\
       	
       	& Norway & Spain 
       	& Schleswig-Holstein & Silesia (\xmark)
       	& Heidelberg & Strasbourg \\
       	
       	& The Netherlands & Denmark 
       	& Rhineland-Palatinate & WestPhalia
       	& Munich & Bonn \\
       	
       	& England & Switzerland 
       	& Mecklenburg-Vorpommern & Saarland 
       	& Cologne & Mannheim \\
       	
  \hline
  \multirow{5}*{\makecell{Separate \\Expansion\\ of Each Set}}      

  		% & Italy & Baden-Wurttemberg   & Strasbourg \\
    %     & Luxembourg 	& WestPhalia    & Berlin \\
       	% & Canada & Hesse    & Marseille \\
       	% & Belgium & Saxony-Anhalt     & Hanover \\
       	% & Spain & Baden     & Auxerre \\
        % & Brussels (\xmark) & Berlin     & Lyon \\
       	% & England & Wurttemberg     & AS Saint-Etienne (\xmark) \\
       	% & Paris (\xmark)  & Munich (\xmark)     & Nancy \\
       	% & Switzerland & Franconia (\xmark)     & Lens \\
       	% & Ireland & Stuttgart (\xmark)     & Paris Saint-Germain (\xmark) \\   
       	
       	& Italy & Luxembourg 
       	& Baden-Wurttemberg & WestPhalia 
       	& Strasbourg & Berlin \\
       	
       	& Canada & Belgium 
       	& Hesse & Saxony-Anhalt 
       	& Marseille & Hanover \\
       	
       	& Spain & Brussels (\xmark)
       	& Baden & Berlin 
       	& Auxerre & Lyon \\
       	
       	& England & Paris (\xmark)
       	& Wurttemberg & Munich (\xmark)
       	& AS Saint-Etienne (\xmark) & Nancy \\
       	
       	& Switzerland & Ireland 
       	& Franconia (\xmark) & Stuttgart (\xmark) 
       	& Paris Saint-Germain (\xmark) & Lens \\
       			
  \hline
\end{tabular}
}
\end{table*}

\vspace{-0.2cm}
\subsection{Parameter Study}
% Since auxiliary sets are updated at each iteration, and from previous results we observe that our model benefits more from auxiliary sets when the ranking list becomes longer. Therefore, we would like to study whether the iteration that we start to generate auxiliary sets would influence the quality of our final output.
Since our model expands $t$ entities into the seed set iteratively, we are interested in how the number of entities expanded at each iteration affects the overall performance of the final output. We vary $t$ from 2 to 50 and plot $MAP@K\ (K=10,20,50)$ on both \textit{Wiki} and \textit{APR} dataset in Figure \ref{fig:parameter}. We observe that the optimal settings for both \textit{Wiki} and \textit{APR} dataset is around $t=5$. The reason is that on one hand, when the number of newly added entities in one iteration is too small, a single noisy term can create a big impact which results in semantic drift. On the other hand, when $t$ is too large, the model benefits minimally from the iterative design since the feature pool is not updated timely.
% there aren't enough entities before selecting context features, so it may be unable to obtain representative context features.

\vspace{-0.2cm}
\subsection{Case Study}

\subsubsection{Qualitative Analysis of Generated Auxiliary Sets}

We are interested in what kind of auxiliary classes would be generated for different semantic classes. In Table \ref{tab:aux} we showcase one auxiliary sets generation result for each semantic class using real queries from the \textit{Wiki} dataset. It is amazing that in most cases, our method can extract meaningful related semantic classes with respect to the target class. For example, a set of \textit{provinces} and \textit{cities} are generated for the category of \textit{countries}. This explains why our model can avoid to wrongly expand more fine-grained location names by comparing them to countries. For the category of \textit{Sports Leagues}, the sports teams affiliated to different sports leagues are extracted, and sports teams can share very similar local context features with sports leagues, such as ``\textit{best player in $\_\_$}''. We can also find \textit{Symptoms} as a new auxiliary set corresponding to the ``\textit{Diseases}'' set. By comparing the difference in their skip-gram features, our multiple sets co-expansion algorithm can distinguish these two classes apart. The high quality and diversity of auxiliary sets and their relations with the core set shows that our framework can be generalized to many other semantic classes.

\vspace{-0.2cm}
\subsubsection{Effectiveness of Multiple Sets Co-Expansion}

 To justify the effectiveness of our multiple sets co-expansion module which takes the auxiliary sets as input to help guide the target set expansion, we randomly select two semantic classes in \textit{Wiki} Dataset and show how the expansion of both target set and auxiliary sets benefit from contrastive feature selection. In Table \ref{satellite} we list 10 entities expanded at the first two iterations using \CoExpan or \CoExpan (no aux.) which is the same as expanding each set separately. 
%  In all three cases, our algorithm successfully extracts meaningful satellite sets that are relevant to the core set but belong to different types. 
%  For the query containing companies, our algorithm outputs some products of companies, and for TV channels it generates some TV shows. There are two satellite sets of the query with countries. One mostly contains states of these countries and the other one mostly contains cities. 
Terms that should not be expanded (since they are in a different category) are marked with (\xmark). 
The result proves the effectiveness of our multiple sets co-expansion module that chooses the most discriminative local context features by comparing seeds from different semantic classes. Specifically, for the query $\{\text{Australia, France, Germany}\}$ that belongs to \textit{Countries}, without guidance of auxiliary sets, the original expansion process can be confused by entities from classes of different granularities (e.g., \textit{Provinces} and \textit{Cities}), as is shown in the table that some cities such as \textit{Brussels} and \textit{Paris}, are expanded.
However, the auxiliary sets generated by our model as shown in both Table \ref{tab:aux} and Table \ref{satellite} are sets of provinces and cities that are located in the seed countries: $\{$Queensland, New South Wales, Saxony$\}$ and $\{$Brisbane, Canberra, Stuttgart$\}$. After scoring the context features by their ability to discriminate different semantic classes, the expansion results of both the target class and the auxiliary sets involve fewer cross-class errors, verifying that the multiple sets co-expansion module can effectively utilize the information provided by auxiliary sets.
\section{Conclusion and Future Work}
In this paper we explore the problem of semantic drift in single query set expansion. Our proposed framework \CoExpan generates auxiliary sets for current expanded results and co-expands the target set and auxiliary sets in an iterative manner. The auxiliary sets generation module retrieves related terms, and then merge the terms by their relations to the expanded seeds, forming related but different semantic classes. The multiple sets co-expansion module later takes both target set and auxiliary sets into consideration and co-expands them by extracting the most discriminative context features. Extensive experiments demonstrate that \CoExpan has a good generalizability by capturing high quality auxiliary sets as rival sets for various queries, and these sets do help guide the expansion of target set to avoid falling into pitfalls of related but different semantic types.

For future work, it is interesting to study how we can leverage auxiliary sets to form different semantic classes in the distributed space, so that not only skip-gram features but also embedding features can be used to discriminate them. Our paper only focuses on named entity expansion, while the problem of expanding generalized nouns (i.e., concepts) is more challenging and worth exploring.

\begin{acks}
Research was sponsored in part by DARPA under Agreements No. W911NF-17-C-0099 and FA8750-19-2-1004, National Science Foundation IIS 16-18481, IIS 17-04532, and IIS-17-41317, and DTRA HDTRA11810026. 
Any opinions, findings, and conclusions or recommendations expressed in this document are those of the author(s) and should not be interpreted as the views of any U.S. Government. The U.S. Government is authorized to reproduce and distribute reprints for Government purposes notwithstanding any copyright notation hereon.
% Research was sponsored in part by U.S. Army Research Lab. under Cooperative Agreement No. W911NF-09-2-0053 (NSCTA), DARPA under Agreements No. W911NF-17-C-0099 and FA8750-19-2-1004, National Science Foundation IIS 16-18481, IIS 17-04532, and IIS 17-41317, DTRA HDTRA11810026, and grant 1U54GM114838 awarded by NIGMS through funds provided by the trans-NIH Big Data to Knowledge (BD2K) initiative (www.bd2k.nih.gov).
% Any opinions, findings, and conclusions or recommendations expressed in this document are those of the author(s) and should not be interpreted as the views of any U.S. Government. The U.S. Government is authorized to reproduce and distribute reprints for Government purposes notwithstanding any copyright notation hereon. 
We thank anonymous reviewers for valuable and insightful feedback.
\end{acks}
\bibliographystyle{ACM-Reference-Format}
\balance
\bibliography{acmart.bib}

%\appendix

\end{document}